\documentclass[10pt,twocolumn,letterpaper]{article}

\usepackage{iccv}
\usepackage{times}
\usepackage{epsfig}
\usepackage{graphicx}
\usepackage{amsmath}
\usepackage{amssymb}

\usepackage{multirow}
\usepackage[inline]{enumitem}
\usepackage{dsfont}
\usepackage[table]{xcolor}
\usepackage{multirow}
\usepackage{pifont}% http://ctan.org/pkg/pifont
\usepackage{algorithm}
\usepackage{algpseudocode}% http://ctan.org/pkg/algorithmicx

\usepackage{caption}
\usepackage{subcaption}
\usepackage{fix2col}
\usepackage{diagbox}
\usepackage{xspace}
\usepackage{tabularx,booktabs}
\usepackage{blindtext}
\usepackage[numbers, sort, compress]{natbib}
\usepackage{self_define}

\usepackage[pagebackref=true,breaklinks=true,colorlinks,bookmarks=false]{hyperref}

\iccvfinalcopy
\def\iccvPaperID{9} % *** Enter the ICCV Paper ID here

\ificcvfinal\fi

\begin{document}

\title{FrankMocap: A Monocular 3D Whole-Body Pose Estimation System \\ 
	            via Regression and Integration}

\author{ Yu Rong\textsuperscript{1,3}\thanks{Work done during Yu Rong's internship at Facebook AI Research}
	\hspace{0.3in} Takaaki Shiratori\textsuperscript{2}
	\hspace{0.3in} Hanbyul Joo\textsuperscript{3}
	\vspace{5pt}
	\\
	\textsuperscript{1}{The Chinese University of Hong Kong}
	\hspace{0.3in} \textsuperscript{2}{Facebook Reality Labs}
	\hspace{0.3in} \textsuperscript{3}{Facebook AI Research} \\ 
}

\let\oldtwocolumn\twocolumn
\renewcommand\twocolumn[1][]{%
	\oldtwocolumn[{#1}{
		\begin{center}
			\includegraphics[width=\linewidth]{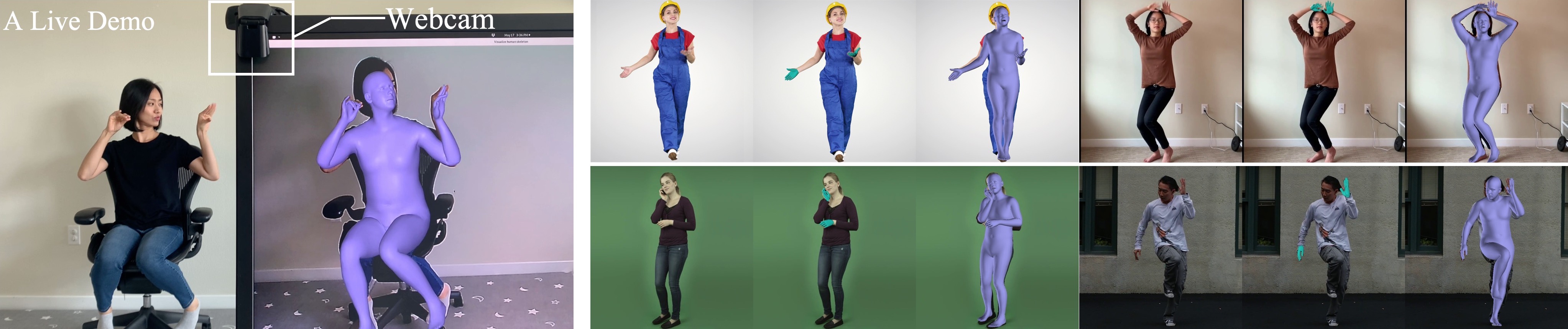}
			\vspace {-0.6 cm}
			\captionof{figure}{\small
				We present \emph{FrankMocap}, a system that estimates 3D poses of face, hands, and body from monocular single images. Our method is fast and capable of performing a live demo with a single RGB webcam, as shown on the left. On the right, example from in-the-wild videos are shown. We show input images (left), 3D hand pose estimation (middle), and the whole-body pose estimation (right).}
				%We present a whole-body 3D pose estimation system, \emph{FrankMocap}, that estimates 3D poses of face, hands, and body from monocular single images. Our method is fast (0.15 sec/frame) and capable of performing a live demo with a single RGB webcam, as shown on the left. On the right, example results from in-the-wild videos are demonstrated. We show input images (left), 3D hand pose estimation outputs from our hand module (middle), and the whole-body pose estimation outputs generated by our system (right).}
			
			%}
			\label{fig:teaser}
		%\vspace{0.0 cm}
		\end{center}
	}]
}

\maketitle
\ificcvfinal\thispagestyle{empty}\fi

% main paper
\begin{abstract}
\vspace{-4px}
%\vspace{-0.26cm}

% 3D human pose estimation facilitate applications such as AR/VR.
%
Most existing monocular 3D pose estimation approaches only focus on a single body part, neglecting the fact that the essential nuance of human motion is conveyed through a concert of subtle movements of face, hands, and body.
In this paper, we present \textrm{FrankMocap}, a fast and accurate whole-body 3D pose estimation system that can produce 3D face, hands, and body simultaneously from in-the-wild monocular images. 
The idea of FrankMocap is its modular design: We first run 3D pose regression methods for face, hands, and body independently, followed by composing the regression outputs via an integration module. 
The separate regression modules allow us to take full advantage of their state-of-the-art performances without compromising the original accuracy and reliability in practice. 
We develop three different integration modules that trade off between latency and accuracy.
All of them are capable of providing simple yet effective solutions to unify the separate outputs into seamless whole-body pose estimation results.
We quantitatively and qualitatively demonstrate that our modularized system outperforms both the optimization-based and end-to-end methods of estimating whole-body pose.
Code and models and demo videos are available at~\url{https://github.com/facebookresearch/frankmocap}.

\end{abstract}
\section{Introduction}
\label{sec:Introduction}

Estimating 3D human pose from single RGB images is one of the core technologies to build a computational model to understand human behavioral cues. 
It can faciliate numerous applications including assistive technology~\cite{gu2019home,leo2018deep,ortega2020dmd}, sign language understanding~\cite{camgoz2020multi}, AR/VR~\cite{bagautdinov2021driving,wang2021scene}, and social signal understanding~\cite{ng2020body2hands}.
Importantly, the essential nuance of human behaviors is conveyed through a concert of subtlest movements of face, hands, and body. 
Thus, it is necessary to estimate whole-body motions to capture the authentic signals.

Estimating whole-body 3D poses including a face and hands, however, remains challenging. 
One of the major difficulties comes from the fact that the scale of faces and hands is much smaller than that of torsos and limbs.
Furthermore, hands in motion are susceptible to artifacts caused by abrupt viewpoint changes, self-occlusions, and motion blur. 
These factors largely raise the difficulties of creating large-scale datasets with whole-body 3D poses, even in a controlled environment~\cite{joo2018total,lee2019talking,pavlakos2019expressive},
not to mention capture such data in the wild.
The lack of whole-body 3D data is a major obstacle preventing the construction of a unified system to estimate whole-body 3D poses simultaneously.
Therefore, most existing methods focus on one of the major body parts individually: single image-based estimation of 3D body pose (\ie, torso and limbs)~\cite{bogo2016keep,kanazawa2018end,xiang2019monocular,kolotouros2019spin,kolotouros2019cmr,wang2020motion,rong2020chasing}, 3D hand pose~\cite{zimmermann2017learning,cai2018weakly,iqbal2018hand, boukhayma20193d,ge20193d,zhang2019end} or 3D face~\cite{tuan2017regressing,chang2018expnet,feng2018joint,sanyal2019learning,cao2018pose}.

%new version
In this paper, we present \textit{FrankMocap}, a modular system for estimating whole-body 3D poses in a unified output in SMPL-X form~\cite{pavlakos2019expressive}, as visualized in Fig.~\ref{fig:teaser}.
Our system is built upon the insight that training a single model to jointly estimate whole parts is intrinsically limited by the lack of accurate and diverse whole-body motion data. 
Instead, we design a modularized system to first run 3D pose regression methods for face, hands, and body independently. 
The three regression outputs are then composed via an integration module. 
The separate regression modules allow us to take full advantage of their state-of-the-art performances without compromising the original accuracy and reliability in practice. 
Our integration module provides a simple yet effective solution to unify them into seamless whole-body pose estimation outputs. 
Unlike previous whole-body pose estimation methods that require computationally heavy optimization~\cite{joo2018total, xiang2019monocular,pavlakos2019expressive}, our method also allows us to run at an interactive frame rate. 
We quantitatively and qualitatively demonstrate that our modular framework outperforms existing optimization-based methods~\cite{xiang2019monocular, pavlakos2019expressive} and end-to-end models trained to produce whole-body outputs jointly~\cite{choutas2020monocular}. 
%
% Code and models will be made publicly available.

In conclusion, we make three major contributions.
First, towards a practical system for whole-body pose estimation, we present the idea of \textit{regression-and-integration} to take full advantage of existing single-part pose datasets and circumvent the hurdle of lacking whole-body pose datasets.
Especially, our hand module is comparable with SOTA hand-only methods on in-the-wild scenarios.
Second, we present three effective strategies to integrate outputs from separate single-part pose estimation modules, namely the fastest ``copy-paste'' approach, a full-optimization approach, and an intermediate approach via a simple integration network.
Third, we quantitatively and qualitatively demonstrate that our modular system outperforms the alternative approaches based on optimization-based methods or end-to-end methods training all parts jointly.

\begin{figure*}[t]
	\centering
	\includegraphics[width=0.95\linewidth]{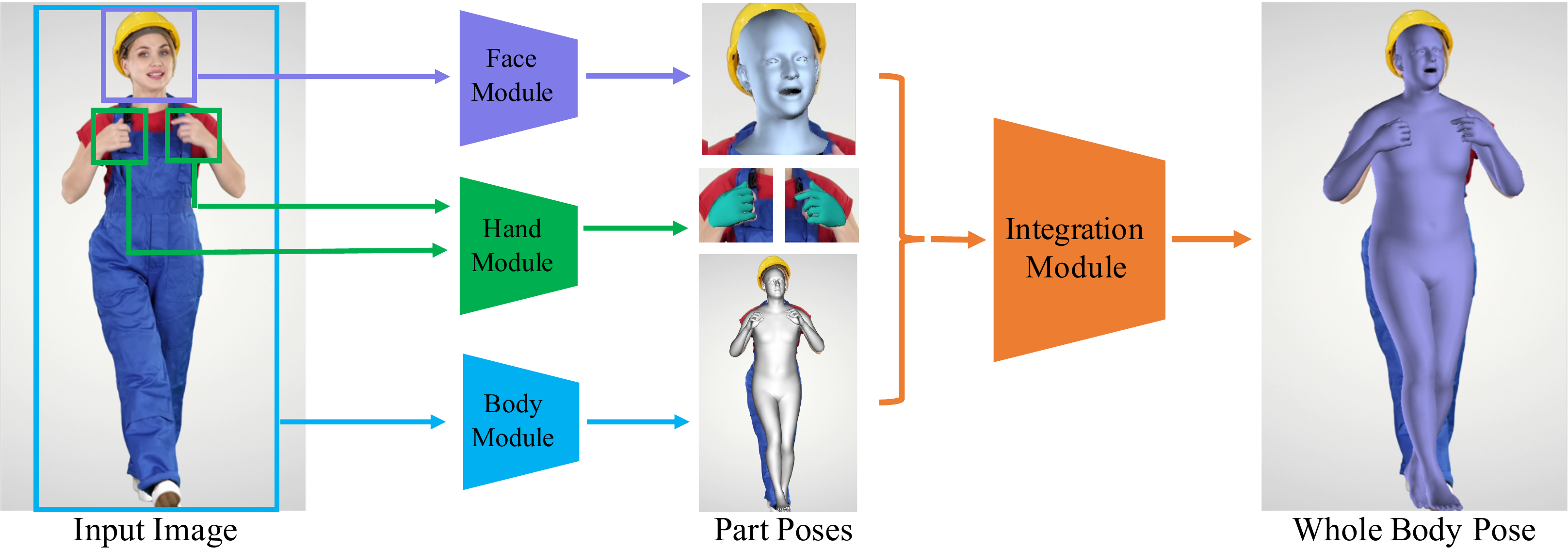}
	\vskip -0.1 cm
	\caption{\small Overview of \textit{FrankMocap}, the proposed whole-body 3D pose estimation system. Given a single RGB image input, we first apply our part modules separately to estimate 3D hands, body, and face. Our integration module is then adopted to combine the outputs from each module into a unified whole-body output.}	\label{fig:overview}
	\vspace{-0.3 cm}
\end{figure*}
\section{Related Work}
\label{sec:RelatedWorks}

\noindent \textbf{3D Parametric Human Body Models.}
3D parametric human models are widely used as a strong prior to estimate 3D pose and shape of humans.
A main idea is to model the deformation of 3D human (including face, hands and body) via low dimension parameters~\cite{anguelov2005scape,pons2015dyna,pavlakos2019expressive,romero2017embodied,joo2018total}. 
SCAPE~\cite{anguelov2005scape} is a pioneering work that accounts for shape variations and pose deformations of human body.
Loper~\etal introduce SMPL~\cite{loper2015smpl} that learns local pose-dependent blendshape on top of linear blend skinning for holistic mesh deformation as well as shape variations.
%
% Later, Romero~\etal~\cite{romero2017embodied} use a similar design to develop the hand deformation model called MANO. 
Similarly, there exists a hand deformation model called MANO~\cite{romero2017embodied} and several face models~\cite{cao2014facewarehouse, blanz1999morphable, li2017learning}.
Recent approaches also introduce whole-body models such as Adam~\cite{joo2018total}, SMPL-X~\cite{pavlakos2019expressive}, and GHUM \& GHUML~\cite{xu2020ghum} that are capable of expressing face, body, and hands in a unified space.

\newpara{Single Image 3D Body, Hand and Face Estimation.}
Recent monocular 3D body motion capture approaches adopt parametric model such as the SMPL~\cite{loper2015smpl} or Adam model~\cite{joo2018total,xiang2019monocular} for 3D body representation\footnote{Note that \textit{body} indicates torso and limbs excluding finger joints and \textit{whole-body} indicates all of face, hands and body.}.
A pioneering work~\cite{bogo2016keep} uses an optimization framework to fit the 3D body model to 2D observations.
More recent methods~\cite{kanazawa2018end,kanazawa2019learning, kolotouros2019spin,tung2017self,omran2018neural,rong2019delving,guler2019holopose,xu2019denserac,moon2020i2l,joo2020exemplar, georgakis2020hierarchical,lin2021end,li2021hybrik} leverage a deep learning framework to regress parameters of the body models from RGB images.
Non-parametric methods have been also introduced by directly regressing the model vertices~\cite{kolotouros2019cmr,choi2020pose2mesh} or corresponding UV maps~\cite{zeng20203d}.
There are also approaches that use a hybrid framework by using a deep learning framework to produce intermediate representations such as 2D heatmaps and then fitting the skeletal models on these outputs to reconstruct joint angles~\cite{mehta2017vnect,xiang2019monocular,mehta2019xnect}.
Due to the lack of in-the-wild training data with 3D annotations, most of these models are trained with mixed datasets, including indoor datasets such as Human3.6M~\cite{ionescu2014human3} and in-the-wild datasets such as COCO~\cite{lin2014microsoft}.
While most methods are based on single images as input, recent approaches~\cite{kanazawa2019learning, kocabas2020vibe} predict 3D motions from a video.
%
% In this paper, we mainly focus on processing single images and applying simple temporal smoothness to handle sequence inputs.

%
Monocular 3D hand pose estimation approaches share a similar pipeline as the body counterparts, based on the parametric 3D hand models, such as MANO~\cite{romero2017embodied}. 
Deep neural networks are leveraged to either predict model parameters~\cite{boukhayma20193d,zhang2019end,baek2019pushing,zhou2020monocular,chen2021model,chen2021camera} or directly regress hand mesh vertices~\cite{ge20193d,kulon2020weakly}.
Similarly, single image 3D face prediction methods~\cite{tuan2017regressing, chang2018expnet, feng2018joint, sanyal2019learning} also share similar ideas by leveraging deep neural networks to regress the face landmarks or parameters of face models, such as 3DMM~\cite{blanz1999morphable} or FLAME~\cite{li2017learning}.
We refer the readers to recent surveys on 3D face reconstruction~\cite{zollhofer2018state,morales2020survey} for more details.

\noindent \textbf{Joint 3D Estimation of Body, Hands and Face.}
There are a few methods~\cite{xiang2019monocular,pavlakos2019expressive,choutas2020monocular,zhou2021monocular} aiming to capture whole-body 3D motions.
SMPLify-X optimizes the parameter of the SMPL-X model~\cite{pavlakos2019expressive} to fit it to 2D keypoints with additional constraints, including body pose priors and collision penalizer.
Monocular Total Capture (MTC)~\cite{xiang2019monocular} is based on the Adam model~\cite{joo2018total}, and adopts deep neural networks to get 2.5D predictions first. Then the parameters of Adam are obtained through optimization.
Both of these methods rely on optimization procedures with relatively slow computational time (from 10 seconds to a few minutes).
Zhou~\etal~\cite{zhou2021monocular} use SMPL-H~\cite{romero2017embodied} to represent body and hands and 3DMM face model~\cite{tewari2017mofa} to represent 3D faces.
They first predict 3D body and hand poses and then separately predict the parameters of 3DMM model, including shape, expression, albedo, and illumination.
The predicted full-body motion is not represented in a unified format and this paper does not present generalization ability to in-the-wild scenarios.
ExPose~\cite{choutas2020monocular}, avoids using the optimization procedure by presenting a neural network to simultaneously predict the body, hand, and face parameters of SMPL-X. 
In particular, ExPose curates a pseudo-ground truth dataset by fitting SMPL-X model on in-the-wild images, followed by manual quality checking by annotators, and use it to train their model to jointly produces output.

\section{Method}
\label{sec:Method}

Given a single image as input, FrankMocap firstly estimates the 3D poses of face, hands (both left and right), and body (the torso and limb parts), in the form of SMPL-X model~\cite{pavlakos2019expressive}. 
Each part's output is produced by a separate 3D pose regressor trained on public datasets in each sub-field. 
Their outputs are combined through our integration module, to produce a seamless and unified whole-body pose estimation output. Fig.~\ref{fig:overview} illustrates an overview of the proposed framework.

\subsection{Whole-Body Parameterization}

\newpara{SMPL-X Model.}
We formulate the SMPL-X model as:

\begin{equation}
\label{eq:smplx}
	\boldsymbol{V}_w = W( \bphi_w, \btheta_w, \bbeta_w, \bpsi_f)\\
\end{equation}

\noindent
where $W$ is parameterized by global orientation of the whole-body $\bphi_w \in \mathbb{R}^{3}$, facial expressions parameters $\bpsi_f \in \mathbb{R}^{10}$,
whole-body pose parameters $\btheta_w  \in \mathbb{R}^{ (21 + 15 + 15 + 1) \times 3}$ accounting for pose-dependent deformation,
and shape parameters $\bbeta_w \in \mathbb{R}^{10}$ accounting for cross-identity shape variations of the face, hands, and body.
We divide $\btheta_w$ for each of the body, hands, and faces, namely
body pose parameters $\btheta_w^b  \in \mathbb{R}^{21 \times 3}$, 
left hand pose parameters $\btheta_w^{lh} \in \mathbb{R}^{15 \times 3}$,
right hand pose parameters $\btheta_w^{rh} \in \mathbb{R}^{15 \times 3}$, 
and face pose parameters\footnote{While the original SMPL-X model has jaw, left and right eye poses for the face part, we do not estimate eye poses and only consider jaw pose.} $\btheta_w^{f} \in \mathbb{R}^{1 \times 3}$.
In this way, $\btheta_w = \{ \btheta_w^b, \btheta_w^{lh}, \btheta_w^{rh}, \btheta_w^{f} \}$.
All pose parameters are defined in the axis-angle representation, which stores the relative rotation to the parent joints defined in the kinematics map. 
As output, the SMPL-X model produces a mesh structure with 10,745 vertices, $\boldsymbol{V}_w \in \mathbb{R}^{10475 \times 3}$. 
The 3D joint locations of the whole-body can be obtained by applying a joint regression function $R$ from the posed vertices as $\boldsymbol{J}^{3D}_{w}  = R_w( \boldsymbol{V}_w)$, where $\boldsymbol{J}^{3D}_{w} \in \mathbb{R}^{ (22 + 20 + 20 + 3) \times 3}$.

\newpara{Stand-Alone Hand Model.}
Our hand model is defined by taking the hand parts of SMPL-X:

\begin{equation}
\label{eq:smplx_hand}
	\boldsymbol{V}_h =  H(\bphi_h, \btheta_h, \bbeta_h), \\
\end{equation}

\noindent
where $\btheta_h \in \mathbb{R}^{3 \times 15}$ is hand pose parameters and $\bbeta_h$ is the shape parameters for the hand model.
Since our hand model is taken from SMPL-X, $\bbeta_h$ shares the same shape space as $\bbeta_w$.
For brevity, we use $\btheta_h$, instead of $\btheta_{rh}$ or $\btheta_{lh}$, to denote the hand pose parameters of either part in our description. 
$\bphi_h \in \mathbb{R}^{3}$ represents the global orientation of the hand meshes.
Our hand model $H$ produces the hand mesh structure with 778 vertices, $\boldsymbol{V}_h \in \mathbb{R}^{778 \times 3}$, where the hand vertices are selected from the hand area of the original SMPL-X vertices. 
Given hand vertices $\boldsymbol{V}_h$, 3D hand joints can be regressed as $\boldsymbol{J}^{3D}_{h}  = R_h(\boldsymbol{V}_h)$, 
where $\boldsymbol{J}^{3D}_{h} \in \mathbb{R}^{21 \times 3}$ contains a wrist, 15 finger joints (3 joints per finger), and 5 finger tips.
Hand visualization and skeleton hierarchy are listed in Appendix~\ref{sec:hand_module_detail}.
The major advantage of our hand representation is that the components of this 3D hand model, including pose parameters, vertices, and 3D joints, are directly compatible with the whole-body parameterization. 
This enables us to efficiently integrate outputs from the body module and the hand module.

\begin{figure}[t]
	\begin{center}
		\includegraphics[width=0.9\linewidth]{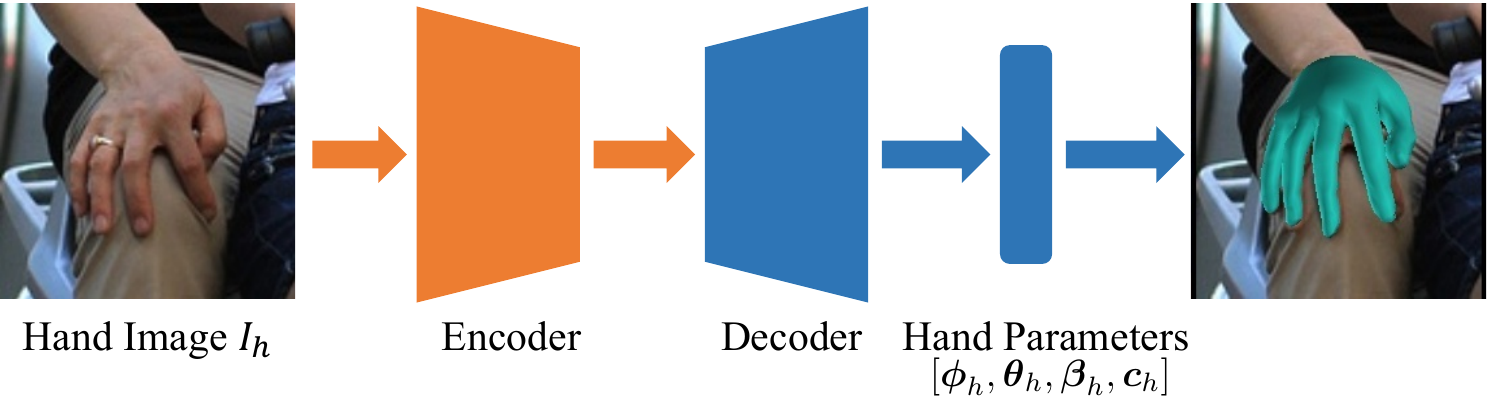}
	\end{center}
	\vskip -0.5 cm
	\caption{\small Overview of hand module. Our hand module takes a cropped hand image $\mathbf{I}_H$ as input, and produces the parameters of hand model, $[\boldsymbol{\phi}_h, \boldsymbol{\theta}_h, \boldsymbol{\beta}_h, \boldsymbol{c}_h] $. The estimated parameters are then used to calculate the mesh of SMPL-X hand part.}
	\label{fig:hand_model}
	\vspace{-0.4 cm}
\end{figure}

\subsection{3D Hand Estimation Module}
\label{sec:hand_module_main}
We present a monocular 3D hand pose estimation module, denoted by $M_H$, to the parameters of the hand model $H$. 
In particular, our hand module is inspired by the recently proposed monocular body pose estimation approaches~\cite{kanazawa2018end, kanazawa2019learning, kolotouros2019spin}. 
We follow similar model architecture, parameterization, and training strategies as the pioneering body methods.
In particular, we introduce blur augmentation to make the hand module more robust to blurry hands that are frequently observed in full body motion capture scenarios.

\noindent \textbf {Network Architecture.} 
Our hand module $M_H$ is built upon an end-to-end deep neural network architecture that can predict hand model parameters defined in Eq.~\eqref{eq:smplx_hand} from given input images. 
This process is defined as:
\begin{equation}
\label{eq:hand_module}
[\bphi_h, \btheta_h, \bbeta_h, \boldsymbol{c}_h]  = M_H(\mathbf{I}_H), \\
\end{equation}
where $\mathbf{I}_H$ is an input RGB image cropped around the hand region. $\boldsymbol{c}_h = (\boldsymbol{t}_h, s_h)$ is a set of weak-perspective camera parameters to project a posed 3D hand model to the input image. 
$\boldsymbol{t}_h \in  \mathbb{R}^{2}$ is for 2D translation on the image plane, and $s_h \in  \mathbb{R}$ is a scale factor. Given camera parameters, the $i$-th 3D hand joint, $\boldsymbol{J}^{3D}_{h,i}$ can be projected as:
\begin{equation}
	\label{eq:2d_projection}
	\boldsymbol{J}^{2D}_{h,i}= s_h \Pi (\boldsymbol{J}^{3D}_{h,i}) +\boldsymbol{t}_h,
\end{equation}
where $\Pi$ is an orthographic projection.
Following the practice of HMR~\cite{kanazawa2018end, kolotouros2019spin}, our hand module $M_H$ adopts an encoder-decoder structure, 
where the encoder (ResNet-50~\cite{he2016deep}) outputs features from an input image, and the decoder regresses the hand model parameters from the features. 
See Fig.~\ref{fig:hand_model} for the overview of our hand module. 
In particular, our hand module is trained with the assumption that the inputs are right hands. 
The images and annotations for the left hand are used after vertical flipping.
During inference, the left-hand images are first flipped and processed as if they were a right hand. Then the outputs are flipped back to the original left-hand space.
Note that the shape parameter $\boldsymbol{\beta}_h$ is originally defined for whole-body model $\boldsymbol{\beta}_w$, but we only consider the deformation for the hand vertices defined in Eq.~\eqref{eq:smplx_hand}, ignoring the body part.

\begin{figure}[t]
	\begin{center}
		\includegraphics[width=1.0\linewidth]{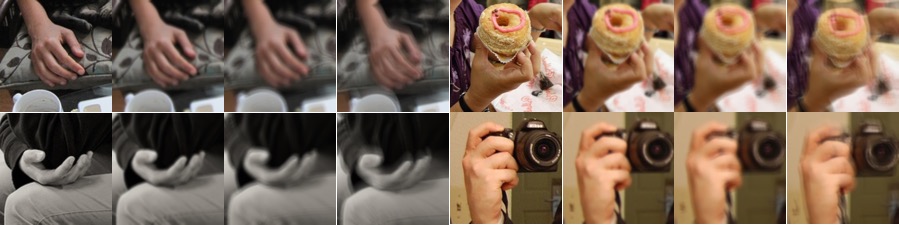}
	\end{center}
	\vskip -0.5 cm
	\caption{\small Motion Blur Augmentation. We show example images of motion blur augmentation. From left to right: original images, augmented images after applying different motion blur kernels.}
	\label{fig:motion_blur}
	\vspace{-0.1 cm}
\end{figure}

\noindent \textbf{Training Method.} 
We consider three different types of annotations: (1) 3D pose annotations (in axis-angle representation), (2) 3D keypoint (joint) annotations, and (3) 2D keypoint annotations. 
The losses for each of the annotation, namely $L_{\btheta}$, $L_{3D}$ and $L_{2D}$, are defined as follows: $L_{\btheta} = \lVert \btheta_h - {\bhtheta}_h \rVert^{2}_{2}$, 
$L_{3D} =  \lVert \boldsymbol{J}^{3D}_{h} - \boldsymbol{\bar{J}}^{3D}_{h} \rVert^{2}_{2}$, and
$L_{2D} = \lVert \boldsymbol{J}^{2D}_{h} -  \boldsymbol{\bar{J}}^{2D}_{h} \rVert_{1}$,
where ${\bhtheta}_h$, $\boldsymbol{\bar{J}}^{3D}_{h}$ and $\boldsymbol{\bar{J}}^{2D}_{h}$ are the ground-truth axis-angle pose parameters, 3D keypoints, and 2D keypoints, respectively.
In particular, the 2D keypoint loss are leveraged to estimate the camera parameters and facilitate the hand module generalize to in-the-wild images with only 2D keypoints annotation.
We do not use the shape parameters provided by the 3D hand datasets such as FreiHAND~\cite{zimmermann2019freihand}, since these are defined for the MANO model~\cite{romero2017embodied} and not compatible with our hand model from SMPL-X.
Instead, an additional shape parameter regularization loss $L_{reg} = \lVert \boldsymbol{\beta}_h \rVert^{2}_{2}$ is applied.
The overall loss $L$ used to train our hand module is defined as follows:
\begin{equation}
\label{eq:overall_loss}
\begin{aligned}
	& L = \lambda_{1}L_{\btheta} + \lambda_{2}L_{3D} + \lambda_{3}L_{2D} + \lambda_{4}L_{reg}.
\end{aligned}
\end{equation}
In experiments, the balanced weights are set as $\lambda_{1} = 10$, $\lambda_{2} = 100$, $\lambda_{3} = 10$ and $\lambda_{4} = 0.1$.
Other hand module training details, including data preprocessing, dataset information, and implementation details, are included in the Appendix~\ref{sec:hand_module_detail}.

\newpara{Motion Blur Augmentation.} 
Performing data augmentations during training is a common practice to enable a model with better generalizability. 
We firstly apply common data augmentation strategies, including random scale, random translation, color jittering, and random rotation. 
Importantly, we recognize that in-the-wild videos are often accompanied by severe motion blur. 
To achieve robustness to motion blur, we additionally apply motion blur augmentation to the images. 
We use the methods~\cite{boracchi2010uniform,boracchi2012modeling} to generate blur kernels and then use 2D filtering to add blurriness to images. 
Examples of applying motion blur augmentation are shown in Fig.~\ref{fig:motion_blur}.
Our experiments clearly show that our motion blur augmentation can effectively insist our hand module generalize to in-the-wild scenarios.

\subsection{3D Body Estimation Module}
Our body module $M_B$ produces the torso and limb parameters defined in Eq.~\eqref{eq:smplx} from a single image:

\begin{equation}
\label{eq:body_inference}
[\boldsymbol{\phi}_b, \boldsymbol{\theta}_b, \boldsymbol{\beta}_b, \boldsymbol{c}_b]  = M_B(\boldsymbol{I}_b), \\
\end{equation}
where $\boldsymbol{I}_b$ is an input image cropped around a target single person's whole-body. 
Similar to Eq.~\eqref{eq:smplx}, $\boldsymbol{\phi}_b \in \mathbb{R}^{3}$ is the global body orientation, $\boldsymbol{\theta}_b  \in \mathbb{R}^{21 \times 3}$ is the body pose parameters (without any finger joints), and $\boldsymbol{\beta}_b \in \mathbb{R}^{10}$ is the shape parameter. $\boldsymbol{\beta}_b$ shares the same parameterization space as $\boldsymbol{\beta}_w$, as defined in Eq.~\eqref{eq:smplx}.
Similar to Eq.~\eqref{eq:hand_module}, we use weak perspective camera parameters  $\boldsymbol{c}_b = (\boldsymbol{t}_b, s_b)$.

We leverage the publicly available body pose estimation model (SPIN~\cite{kolotouros2019spin}) trained by the EFT dataset~\cite{joo2020exemplar}. 
Since the original SPIN model is built with SMPL, we fine-tune the network by replacing SMPL with SMPL-X.
For the fine-tuning process, we use the existing indoor 3D pose dataset Human3.6M~\cite{ionescu2014human3} and outdoor pseudo-GT dataset by EFT~\cite{joo2020exemplar} that provides SMPL annotations.
Since the annotation (SMPL) and our target model (SMPL-X) are not exactly compatible, we only use the pose parameters and 2D keypoint from the annotations during fine-tuning process, ignoring the shape parameters
\footnote{Even if the 3D joint locations of SMPL and SMPL-X are not exactly identical, we found that this distinction is not very significant in practice.}.

\subsection{3D Face Estimation Module}
We use the off-the-shelf public 3D face estimation model RingNet~\cite{sanyal2019learning} to estimate facial expressions $\boldsymbol{\psi}_f$ and face poses $\boldsymbol{\theta}_f$ as follows:
\begin{equation}
\label{eq:face_inference}
[\boldsymbol{\theta}_f, \boldsymbol{\psi}_f]  = M_F(\boldsymbol{I}_f). \\
\end{equation}
Since RingNet is based on FLAME~\cite{li2017learning}, its predictions are compatible with the face part of SMPL-X.
As an adjustment, we only keep the first $10$ expression parameters from the original $50$ dimensional expression parameters $\hat{\psi_f}$ predicted by RingNet, to make it compatible with the expression space defined in SMPL-X.
It is worth noting that the FLAME and SMPL-X share the same PCA-based expression space, while SMPL-X only uses the $10$ foremost principal components.

\subsection{Whole-Body Integration Module}
\label{sec:integration_main}
Our integration module combines the outputs from the face, hands, and body modules into a unified representation of the SMPL-X~\cite{pavlakos2019expressive}. 
While the output of the face module can be easily applied to the face part of the SMPL-X model by copying the expression parameters and jaw poses,
additional procedures are required to integrate the outputs of the hand and body modules because different wrist poses are estimated from these two modules. 
For this objective, we present three strategies: (1) by simple copy-paste composition for fastest processing, (2) by an optimization framework, (3) by an integration network to approximate the optimization via a simple neural network.

\noindent \textbf{By Copy-Paste.}
Our hand and body modules' outputs can be efficiently combined, since they are both defined in SMPL-X parameterization.
As the simplest strategy, we transfer the corresponding joint angle parameters from the outputs of each module into the whole-body model. 
The wrist pose parameters require additional processing because we obtain two different outputs from the body module and the hand module (the wrist pose of hand module is represented by the global hand orientation $\bphi_h$). 
Let us denote the pose parameters for the wrist joint as $\btheta^{\text{wrist}}$, then $\btheta_b = \hat{\btheta}_b \cup \{  \btheta_b^{\text{rwrist}}, \btheta_b^{\text{lwrist}} \} $, where $\hat{\btheta}_b$ represents all body pose parameters except wrists. 
We use the similar notations for the whole-body pose parameters, $\btheta_w^b = \hat{\btheta}_w \cup \{  \btheta_w^{\text{rwrist}}, \btheta_w^{\text{lwrist}} \} $. Then, whole-body integration by copy-paste can be performed as: 
\begin{equation}
\label{eq:copy_paste}
\begin{aligned}
	\left( \bphi_w, \bbeta_w, \boldsymbol{c}_w, \hat{\btheta}_w^b, \btheta_w^{lh}, \btheta_w^{rh} \right) & 
	= \left(  \bphi_b,\bbeta_b , \boldsymbol{c}_b, \hat{\btheta}_b, \btheta_{lh}, \btheta_{rh} \right), \\
	\left( \btheta_w^{\text{lwrist}}, \btheta_w^{\text{rwrist}} \right) & = \left( \Gamma_l \left( \btheta_b, \bphi_{lh}   \right), \Gamma_r \left( \btheta_b, \bphi_{rh}   \right)  \right), \\
\end{aligned}
\end{equation}
where $\Gamma_{\{l,r\}}$ are the functions to convert the global wrist orientation $\bphi_{h}$ obtained from the hand module to the local wrist pose parameters w.r.t. its parent joint in the SMPL-X skeleton hierarchy. 
This can be implemented by comparing $\bphi_{h}$ with the global orientation of the current wrist pose from $\btheta_b$ that can be computed by following the forward kinematics of the body skeleton hierarchy.
This strategy requires almost no extra computation, making our separate modules contribute a common whole-body model simultaneously. 
We found this simple integration produces convincing results, especially for the scenarios with computational bottlenecks as in our live demo.

\begin{figure}
	\centering
	\includegraphics[width=\linewidth]{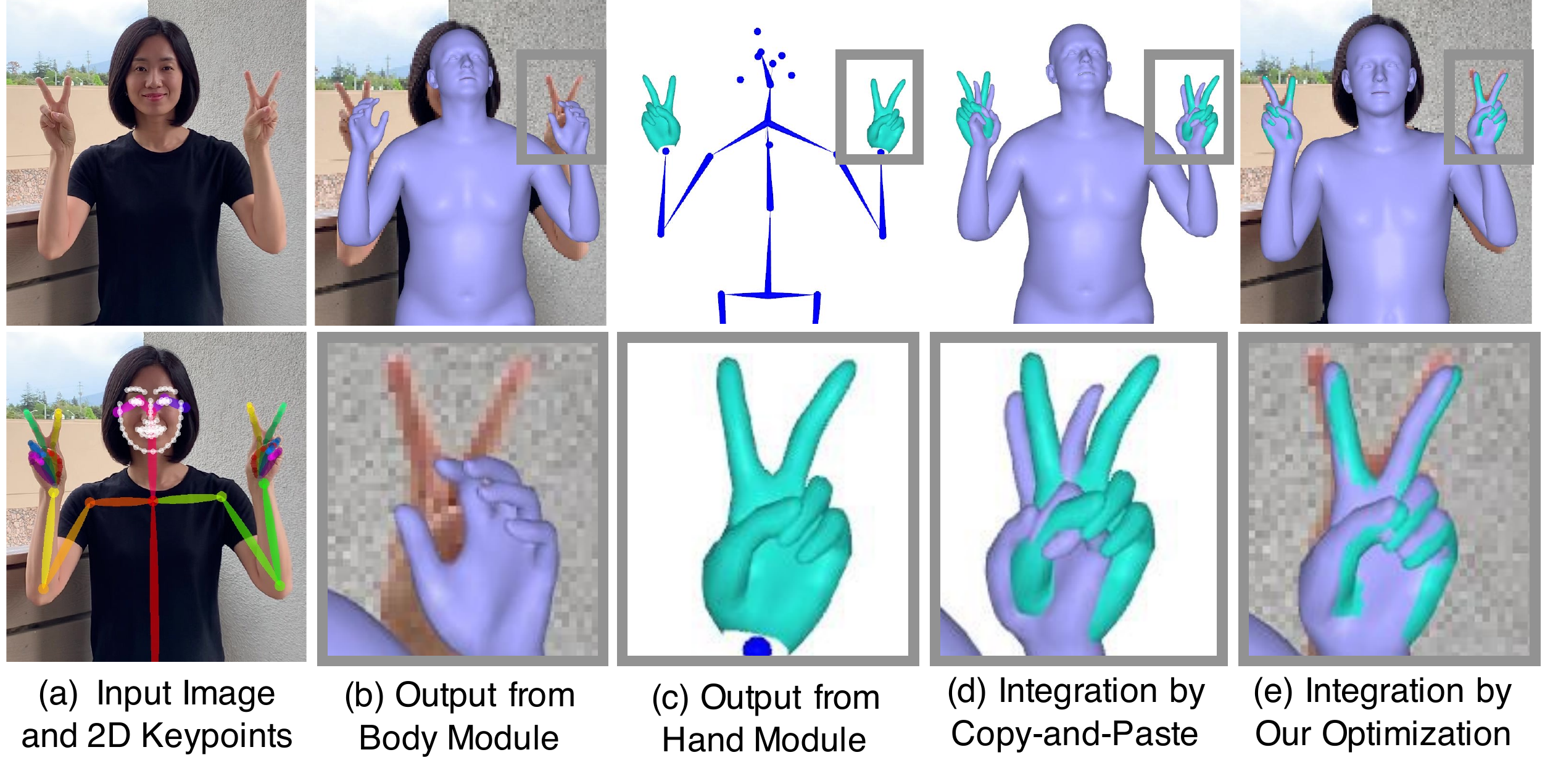}
	\vskip -0.2cm
	\caption{\small Optimizing the whole-body model (SMPL-X) with 3D hand prediction and 2D keypoint estimation. (a) Input images and estimated 2D keypoints by OpenPose~\cite{cao2019openpose}; (b) 3D body pose estimation from body module; (c) The output of 3D hand module aligned to the wrist joints of SMPL-X; (d) Integration output by copy-paste ; (e) Integration output by optimization framework.}
	\label{figure:wholebody_optimization}
	\vspace {-0.2cm}
\end{figure}

\noindent \textbf{By Optimization.}
As an alternative integration method for better accuracy, we build an optimization framework to fit the whole-body model parameters given the outputs from body and hand modules. 
This strategy is particularly helpful to reduce the artifact around the wrist parts over the copy-paste strategy. 
It can also take advantage from the 2D keypoint estimation output~\cite{cao2019openpose} for better 2D localization quality. 
Our optimization framework finds the whole-body model parameters that minimize the following objective cost function:
\begin{equation}
	\label{eq:full_opt}
    \begin{gathered}
        \mathcal{F}([\bphi_w, \btheta_w,   \bbeta_w, \boldsymbol{c}_w]) = \mathcal{F}^{2d} + \mathcal{F}^{mesh} + \mathcal{F}^{pri},
    \end{gathered}
\end{equation}
where $\mathcal{F}^{2d}$ is the 2D re-projection cost term between the 2D keypoint estimation~\cite{cao2019openpose} and the projection of 3D joints (hand, body and faces).
$\mathcal{F}^{mesh}$ is the 3D distance loss between the hand mesh of whole-body model and the mesh vertices of the hand module's output localized on the body's wrist joints, as shown in Fig.~\ref{figure:wholebody_optimization}. 
The prior term $\mathcal{F}^{pri}$ is needed to keep the 3D pose and shape parameters in plausible space.
We use the EFT introduced in \cite{joo2020exemplar} for the similar goal, by applying neural network finetuning to replace explicit 3D pose prior terms used by previous works~\cite{pavlakos2019expressive}.
Therefore, the prior term only accounts for the shape regularization $\mathcal{F}^{pri}=\| \bbeta_w \|^2_2$.

We use a multi-stage approach in which optimization is firstly performed without $\mathcal{F}^{mesh}$.
Then we optimize all terms in the second stage.
During the second stage, we add an additional constraint $\mathcal{F}^{3D}$ to maintain the 3D joints within the final locations of the first stage.
Our experiments show that this term can produce reliable results in practice.
See Fig.~\ref{figure:wholebody_optimization} for the example of our optimization.

\begin{figure}[t]
	\begin{center}
		\includegraphics[width=1.0\linewidth]{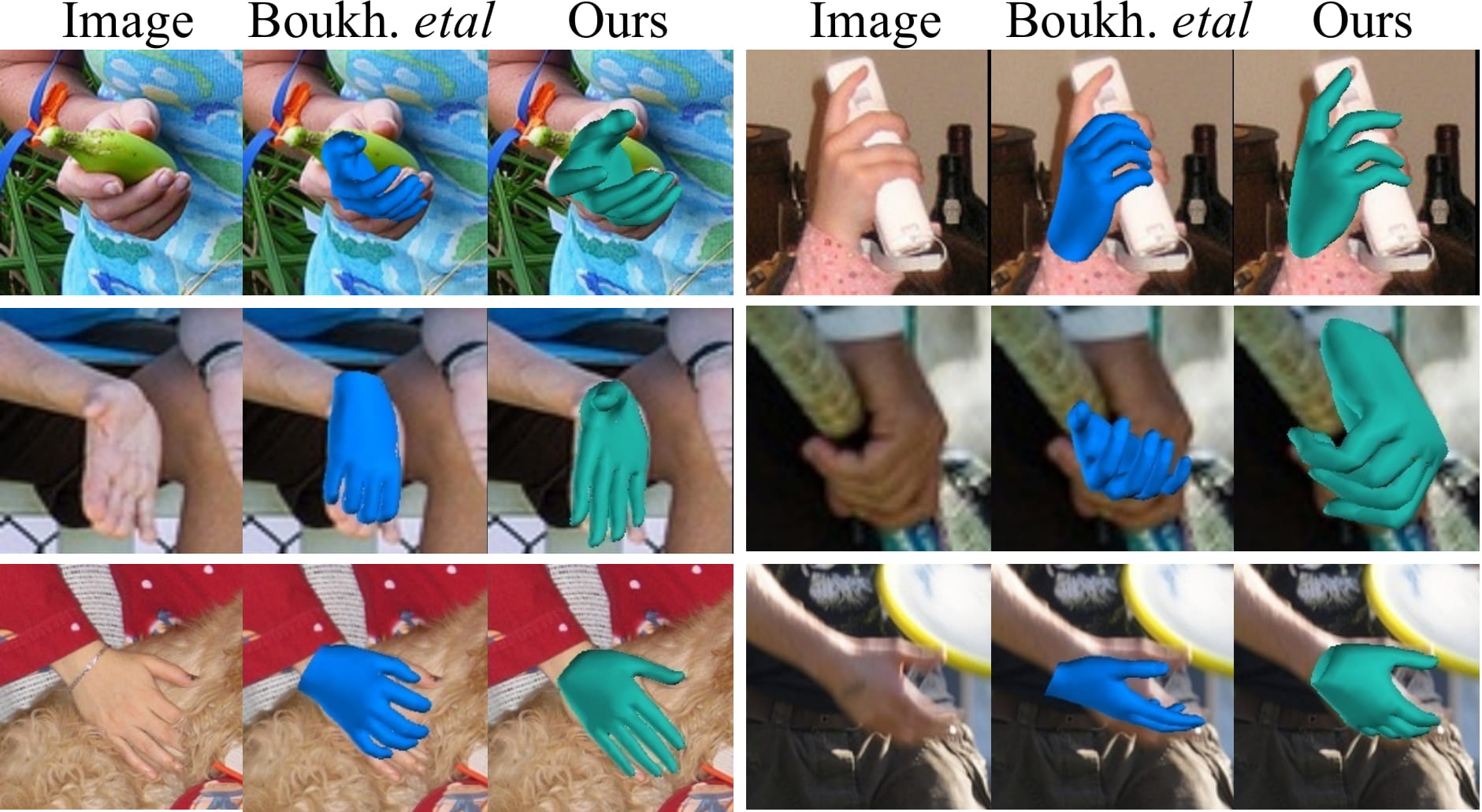}
	\end{center}
	\vskip -0.6 cm
	\caption{\small Qualitative comparison with state-of-the-art 3D hand methods. Images are selected from COCO~\cite{lin2014microsoft}.}
	\label{fig:compare_sota_hand}
	\vspace{-0.3cm}
\end{figure}

\newpara{By Wrist Integration Network.}
The result of the hand module shows precise 2D localization quality as shown in Fig.~\ref{fig:compare_sota_hand}. 
But we lose its precision by applying copy-paste integration strategy since the wrist joint location is determined by the body module, as shown in Fig.~\ref{figure:wholebody_optimization} (second last column).
Although the optimization method can achieve better precision and resolve this problem, it suffers from relatively slow computation speed due to the requirement of iterative gradient computations. 
As another alternative strategy to achieve both precision and fast-runtime performance, we introduce an integration network to adjust the arm pose of the whole-body model toward the given 2D location without gradient computation. 
Specifically, we develop a small neural network to achieve this:
\begin{equation}
    \begin{gathered}
        \tilde{\btheta}_w^{arm} = \mathcal{H}(\btheta_w^{arm}, \textbf{d}),
    \end{gathered}
	\label{eq:wrist_integration}
\end{equation}
where $\textbf{d} \in \mathbb{R}^{2}$ is a 2D directional vector that the arm needs to follow in the image space.
It is obtained from the wrist locations calculated from the body module and those from the hand module, followed by a normalization with the length of the arms.
$\btheta_w^{arm}$ is the pose parameters of the elbow and shoulder joints after the copy-paste integration. 
Conceptually, $\mathcal{H}$ predicts an approximation of optimization gradient to adjust the arm parameters, similar to the recent work of Song~\etal~\cite{song2020human}.
We use a MLP with six layers to implement $\mathcal{H}$.
To train $\mathcal{H}$, we generate a synthetic dataset by capturing our own a range of motion videos to cover diverse arm pose variations and by simulating arbitrary $\textbf{d}$ directions. 
More details of wrist integration are described in Appendix~\ref{sec:wrist_integration_supp}.

\section{Experiments} \label{sec:Results}

\begin{table}[t] \centering \small
	%\rowcolors{3}{lightgray}{}
	\caption{\small Comparison of our hand module with the state-of-the-art 3D hand methods using 2D/3D AUC of PCK.}
	\vskip -0.3 cm
	\setlength\tabcolsep{2pt}
	\resizebox{\columnwidth}{!}{
		\begin{tabular}{c|c|c|c|c|c|c|c}
			
			\toprule
			Method $\rightarrow$  & \multirow{2}{*}{Bouk~\cite{boukhayma20193d}} & \multirow{2}{*}{Baek~\cite{baek2019pushing}} & \multirow{2}{*}{Ge~\cite{ge20193d}} & \multirow{2}{*}{Zhang~\cite{zhang2019end}} & \multirow{2}{*}{Kulon~\cite{kulon2020weakly}} & \multirow{2}{*}{CMR~\cite{chen2021camera}} & \multirow{2}{*}{Ours} \\
			Dataset $\downarrow$ & & & & & & & \\
			\midrule  
			STB 				& 0.994 &   -  & \textbf{0.995} & \textbf{0.995}    & -		& -		 & 0.992  	\\
			RHD			        &  -     & 0.926  & 0.92  & 0.901 & \textbf{0.95}  & 0.944 & 0.934   \\
			MPII+NZSL		    & 0.501 &   - & 0.15 & - & \textbf{0.701} & -     & 0.655  \\
			\bottomrule	
		\end{tabular}	
	}
	\label{tab:hand_compare_with_sota}
	\vspace{-0.2 cm}
\end{table}

\begin{table}[t] \centering \small %%\scriptsize
	%\rowcolors{1}{}{lightgray}
	%\setlength\tabcolsep{2pt}
	\caption{\small Ablation study on hand module data augmentation. Models are evaluated on MPII+NZSL~\cite{simon2017hand} with 2D AUC.}
	\vskip -0.3cm
	\resizebox{\columnwidth}{!}{
		\begin{tabular}{c|c|c|c|c||c}
			\toprule 
			Translation & Rescale & Color & Rotation & Blur & 2D AUC\\
			\midrule  
			\xmark       & \xmark    & \xmark             & \xmark       & \xmark      & 0.608            \\
			\cmark       & \cmark    &                    &              &             & 0.610 \\
			\cmark       & \cmark    & \cmark             &              &             & 0.618 \\
			\cmark       & \cmark    & \cmark             & \cmark       &             & 0.622 \\
			\cmark       & \cmark    & \cmark             & \cmark       & \cmark      & 0.655 \\
			\bottomrule
		\end{tabular}	
	}
	\vspace{-0.35 cm}
	
	\label{tab:ablation_data_augmentation}
\end{table}

\begin{figure}[t]
	\begin{center}
		\includegraphics[width=0.95\linewidth]{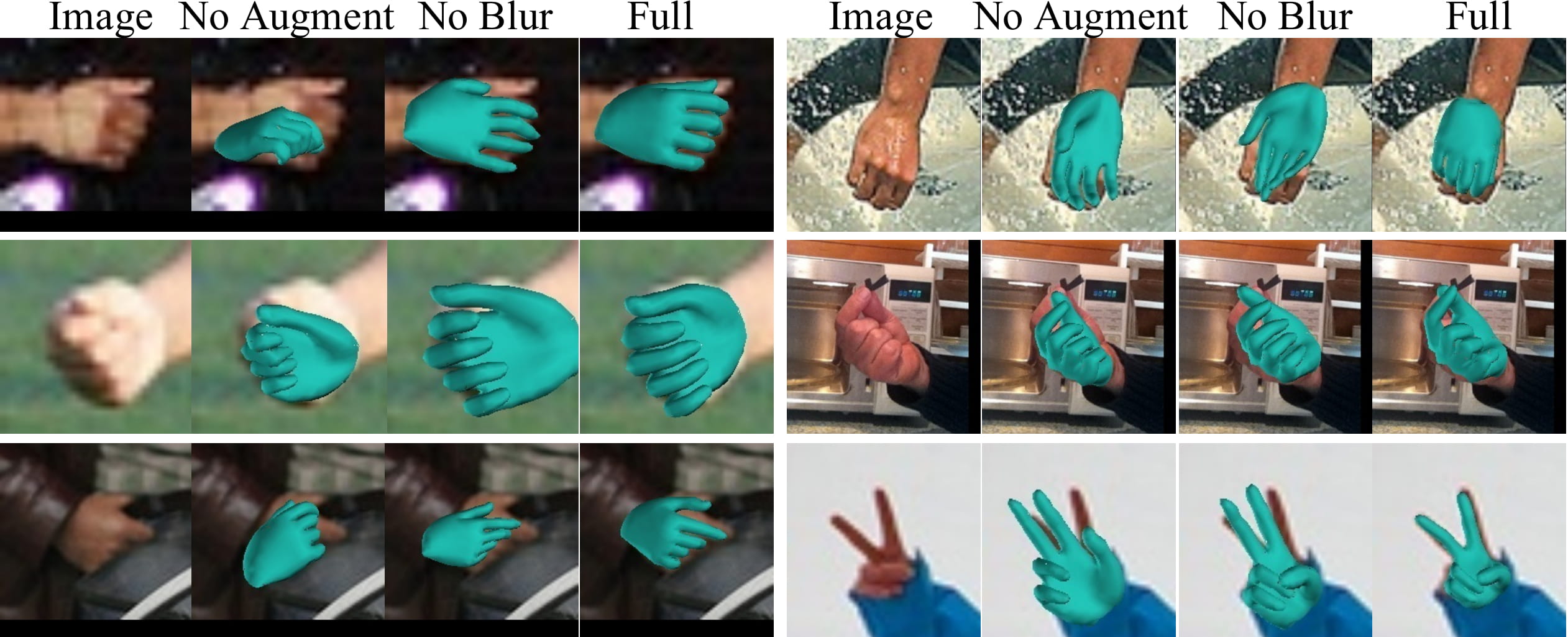}
	\end{center}
	\vskip -0.6 cm
	\caption{\small Ablation study on data augmentations.}
	\label{fig:ablation_augment}
	\vspace{-0.5cm}
\end{figure}

We first demonstrate that our hand module achieves high performance competing to the previous state-of-the-art 3D hand methods, along with ablation studies to examine the designs of our method. 
Then, we demonstrate that our whole-body pose estimation method outperforms previous approaches on a public benchmark. See Appendix~\ref{sec:hand_module} for the details of implementation and datasets.

\subsection{Hand Module Evaluation}

\newpara{Comparison with State-of-the-art Methods.}
We compare our hand module with the previous state-of-the-art hand approaches on three public hand benchmarks, namely STB~\cite{zhang20173d}, RHD~\cite{zimmermann2017learning} and MPII+NZSL~\cite{simon2017hand}. 
The results are listed in Tab.~\ref{tab:hand_compare_with_sota} and Fig.~\ref{fig:compare_sota_hand}. 
For each validation dataset, we calculate the percentage of correct keypoints (PCK) under different thresholds and calculate the corresponding Area Under Curve (AUC) for PCK. 
For STB~\cite{zhang20173d} and RHD~\cite{zimmermann2017learning}, we use 3D AUC and the threshold ranges from $20$ mm to $50$ mm. 
For MPII+NZSL~\cite{simon2017hand}, we use 2D AUC and the threshold ranges from $0$ px to $30$ px.
For fair comparison, all the methods take a single RGB image as input.
As shown in the results, our hand module is comparable with the previous state-of-the-art hand-only methods based on MANO~\cite{romero2017embodied}, though our hand model, taken from the whole-body model, has restricted shape variations
\footnote{Note that the code of Kulton~\etal~\cite{kulon2020weakly} is not publicly available, and, thus, we are unable to make more comparisons in challenging in-the-wild scenes, our target applications.
CMR~\cite{chen2021camera} requires camera intrinsic as known, which prevents applying them to in-the-wild scenarios.}.

\newpara{Ablation Study.}
We examine the data augmentation strategies used in training hand module.
The results are listed in Tab.~\ref{tab:ablation_data_augmentation} and Fig.~\ref{fig:ablation_augment}.
The results in Tab.~\ref{tab:ablation_data_augmentation} demonstrate that applying data augmentation leads to better results. 
The qualitative results in Fig.~\ref{fig:ablation_augment} further show that adopting data augmentation helps our model generalize better on challenging scenarios, including blur, pose variations, and occlusion.
In Fig.~\ref{fig:ablation_augment}, ``No Augment'' refers to the model trained without any data augmentation, and ``No Blur'' refers to the model trained with all data augmentation strategies except the motion blur augmentation.
 ``Full'' refers to the model trained with all data augmentation strategies. 
We also do an ablation study on mixture usage of datasets, demonstrating that using diverse datasets improves the performances. 
See Appendix~\ref{sec:ablation_hand_dataset} for details on this part.

\subsection{Body Module Evaluation}

\begin{table}[t]
	\centering %
	%\small
	%\rowcolors{3}{lightgray}{}
	\caption{\small Comparison of our body module with the state-of-the-art body-only methods on 3DPW~\cite{vonMarcard2018}.} %The evaluation metrics are MPJPE and PA-MPJPE.}
	\vskip -0.3 cm
	\setlength\tabcolsep{2pt}
	\resizebox{\columnwidth}{!}{
		\begin{tabular}{c|c|c|c|c ? c|c}
			\toprule
			Method $\rightarrow$  & \multirow{2}{*}{HMR~\cite{kanazawa2018end}} & \multirow{2}{*}{CMR~\cite{kolotouros2019cmr}} & \multirow{2}{*}{SPIN~\cite{kolotouros2019spin}} & %\multirow{2}{*}{I2L~\cite{moon2020i2l}} &
			\multirow{2}{*}{EFT~\cite{joo2020exemplar}} &
			\multirow{2}{*}{ExPose~\cite{choutas2020monocular}} 
			& \multirow{2}{*}{Ours} \\
			Metric $\downarrow$ & & & & & & \\
			\midrule  
			MPJPE	 & 130                & 127.2               &  96.9				 				&   \textbf{92.3} & 93.4 	&   94.3 \\
			PA-MPJPE & 81.3 		    &  70.2              &  59.2 				 			&  \textbf{54.2} &  60.7 			&  60.0 \\
			\bottomrule	
		\end{tabular}	
	}
	\label{tab:body_compare_with_sota}
	\vspace{-0.5 cm}
\end{table}

We compare our body module with the previous SOTA body-only methods on the test set of 3DPW~\cite{vonMarcard2018}. 
Following the previous approaches, we use Mean Per Joint Position Errors (MPJPE) and PA-MPJPE (after Procrustes analysis~\cite{gower1975generalized}). 
%$
The results are listed in Tab.~\ref{tab:body_compare_with_sota}. 
Our body module is comparable to the previous state-of-the-arts, including ExPose~\cite{choutas2020monocular}. 
Our body model is a fine-tuned version of SPIN~\cite{kolotouros2019spin} and EFT~\cite{joo2020exemplar}.
The slightly degraded performance might be due to the reason that we only use fewer datasets, COCO and Human3.6M datasets. 
We only compare with HMR~\cite{kanazawa2018end}-based methods.
We believe that the experimental results shown in Tab.~\ref{tab:body_compare_with_sota} can demonsrate the feasibility of extending previous SMPL-based methods to SMPL-X based body module.
Thanks to our modularized design, we can also easily leverage the most recent 3D body poses methods~\cite{lin2021end,li2021hybrik} to build our body module.

\subsection{Integration Module Evaluation}

\begin{table}[t] 
	\begin{center}
		\caption{\small Quantitative comparison with previous SMPL-X based whole-body model on EHF dataset~\cite{pavlakos2019expressive}.} %We also compare runtime of diffrent models.}
		\vskip -0.3cm
		\setlength\tabcolsep{2pt}
		\resizebox{\columnwidth}{!}{
			\begin{tabular}{c|c|cc|ccc}
				\toprule	
				\multicolumn{2}{c|}{Methods $\rightarrow$}				& \multirow{2}{*}{SMPLify-X~\cite{pavlakos2019expressive}} & \multirow{2}{*}{ExPose~\cite{choutas2020monocular}} & \multirow{2}{*}{FM-CP} & \multirow{2}{*}{FM-WI} &\multirow{2}{*}{FM-OPT} \\
				\multicolumn{2}{c|}{Part \& Metrics $\downarrow$}			& & & & & \\
				\midrule 
				\multirow{2}{*}{All} 		& V2V			 &	146.2 & 81.9  & 69.9  & 69.3 & \textbf{63.5} \\
				                            & PA-V2V		 &  68.0 & 57.4  & 58.2  & 57.1 & \textbf{54.7} \\
				\midrule
				\multirow{2}{*}{Body} 		& V2V			 &	 231.3 & 116.4 & \textbf{83.7} & 83.9 & 93.1 \\
				                            & PA-V2V		 &  75.4 & 55.1  &  \textbf{52.7} & 53.2 & 53.3 \\
				\midrule
				\multirow{2}{*}{L Hand} 	& V2V			  &	 34.6 & 39.3  & 27.1  & \textbf{27.0} & 28.6 \\
				                            & PA-V2V		  &  11.4 & 13.1  & 10.9  & 10.9 & \textbf{10.8} \\
				\midrule
				\multirow{2}{*}{R Hand}		& V2V			  &	 34.3 & 39.3  & \textbf{25.4}  & 25.5 & 25.6 \\
				                            & PA-V2V		  &  13.1 & 12.9  & \textbf{11.2}  & \textbf{11.2} & \textbf{11.2} \\				  
				\midrule
				\multirow{2}{*}{Face}       & V2V			  & 23.6  & \textbf{19.5}  & 30.7  & 30.7 & 20.4 \\
				                            & PA-V2V		  & 8.4   & 5.6   & 5.7   & 5.7  & \textbf{5.4} \\  
				\midrule
				\multicolumn{2}{c|}{Runtime (sec/frame)} & 60 & 0.15 & 0.15 & 0.15 & 2 \\
				\bottomrule
			\end{tabular}
		}
		\label{tab:compare_sota_whole}
	\end{center}
	\vspace{-0.7cm}
\end{table}

\begin{figure}[t]
	\begin{center}
		\includegraphics[width=\linewidth]{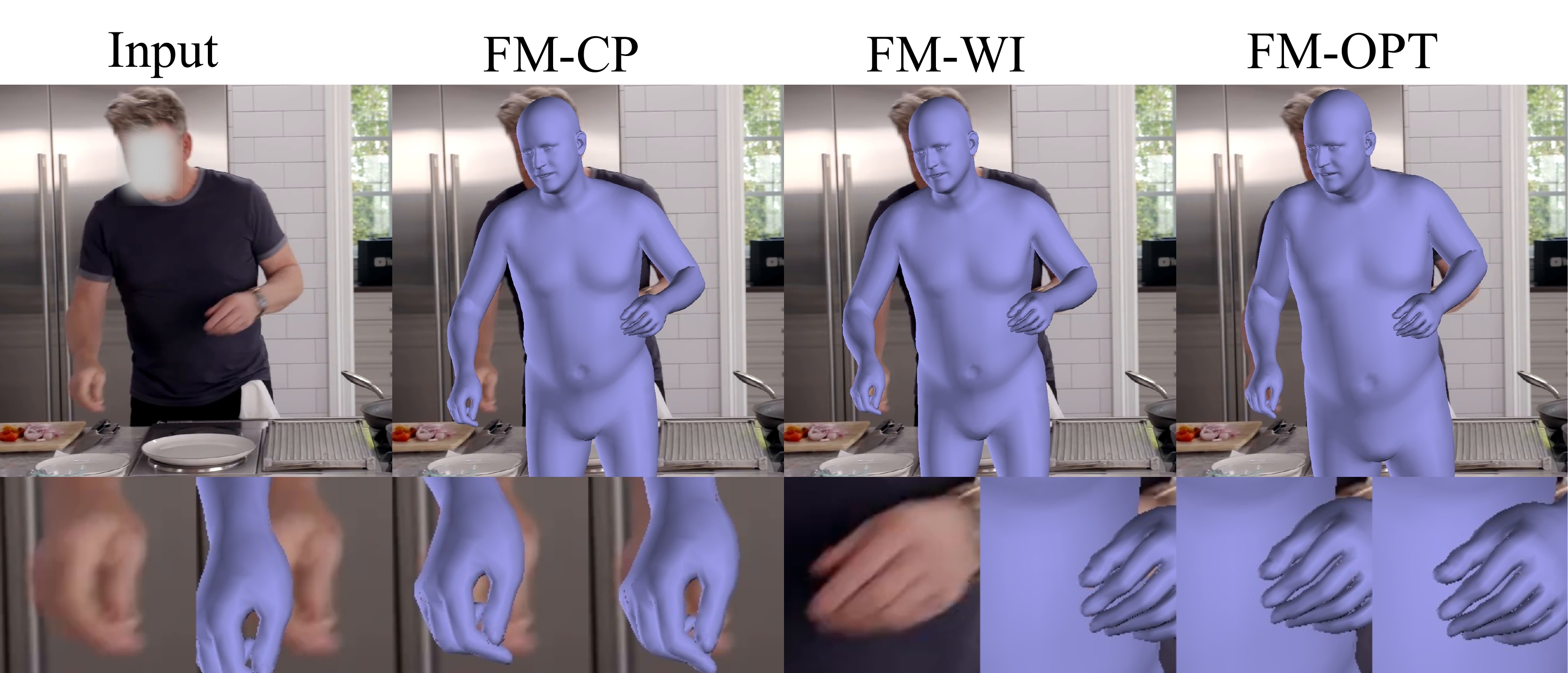}
	\end{center}
	\vskip -0.5 cm
	\caption{\small Qualitative comparison of FrankMocap model with different integration modules.}
	\label{fig:fm_ablation}
	\vspace{-0.4 cm}
\end{figure}

\begin{figure*}[t]
	\begin{center}
		\includegraphics[width=0.95\linewidth]{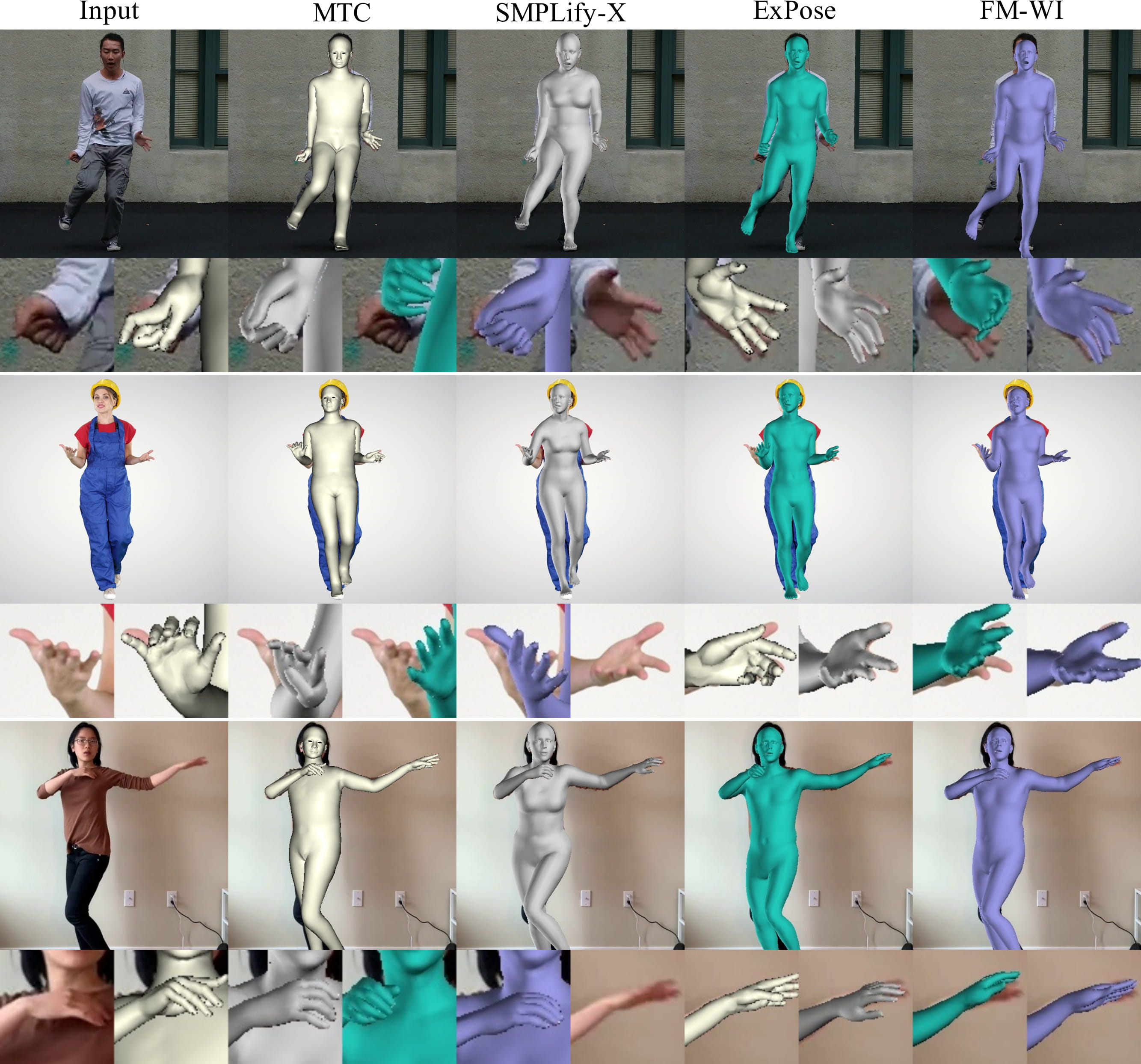}
	\end{center}
	\vskip -0.5 cm
	\caption{\small Qualitative comparison over the previous whole-body pose estimation methods. We compare our method adopting the wrist integration module (FM-WI) with MTC~\cite{xiang2019monocular}, SMPLify-X~\cite{pavlakos2019expressive} and ExPose~\cite{choutas2020monocular}.} %Our method is very efficient, yet producing whole-body poses that aligns well to the actual pose of the person.}
	\label{fig:compare_sota_whole}
	\vspace{-0.4 cm}
\end{figure*}

We quantitatively compare our methods with previous SMPL-X based whole-body methods (SMPLify-X~\cite{pavlakos2019expressive} and ExPose~\cite{choutas2020monocular}) on EHF dataset~\cite{pavlakos2019expressive}. 
The evaluation metrics are vertex-to-vertex distance (V2V) and Procrustes Analysis vertex-to-vertex distance (PA-V2V). 
The results are listed in Tab.~\ref{tab:compare_sota_whole}.
We run the officially released codes of SMPLify-X and ExPose to obtain the evaluation results.
FM-CP, FM-WI, and FM-OPT indicate FrankMocap using the copy-paste, wrist integration network, and optimization-based as the integration module, respectively.
The results demonstrate that all our methods significantly outperform the previous methods in terms of whole-body estimation, body estimation, and hand estimation. 
Our face estimation is comparable with previous methods, where we only use an off-the-shelf 3D face model~\cite{sanyal2019learning}.
Comparing across different versions of FrankMocap reveal that the model with optimization achieves the best performance in general. 
Applying our wrist integration network can also lead to better performance than copy-paste version.
We also provide qualitative comparisons between different integration modules in Fig.~\ref{fig:fm_ablation}. 
It is revealed that using the wrist integration network can bring better hand location while adopting optimization can further improve hand localization while updating body shape and other poses.

We also qualitatively compare our method with previous whole-body motion capture approaches, MTC~\cite{xiang2019monocular}, SMPLify-X~\cite{pavlakos2019expressive}, and ExPose~\cite{choutas2020monocular}, with our output with wrist integration network. 
Results shown in Fig.~\ref{fig:compare_sota_whole} and the demo video demonstrate that our method outperforms previous approaches, with fast and accurate performance in challenging in-the-wild scenes.
Qualitative comparison with Zhou~\etal~\cite{zhou2021monocular} is included in the Appendix~\ref{sec:qualitative_supp}.

\vspace {-0.2 cm}
\section{Conclusion}
\label{sec:Conclusion}
\vspace{-0.1cm}

We present FrankMocap, a monocular 3D whole-body motion capture system built upon the motivation to take full advantages of state-of-the-art part estimation modules, followed by effective integration algorithms.
We develop three integration modules, namely copy-paste, wrist integration network, and an optimization-based method.
The copy-paste and wrist integration modules are suitable for time-sensitive applications, while the optimization-based module is suitable for offline applications requiring better precision. 
Our system surpasses previous state-of-the-art 3D whole-body estimation methods on both public benchmarks and in-the-wild scenarios.
Notably, our approach can directly leverage future algorithms by replacing each module, which is a main advantage of our modular design.

\vspace {0.15cm}
\noindent \textbf{Acknowledgements.} We acknowledge Yuting Ye from FRL, Donglai Xiang and Yu Ding from CMU, Yu Xiong, Anyi Rao, and Jingbo Wang from CUHK, and Jie Chen from Microsoft.
%}

\clearpage
% !TEX root = ../main.tex

\section*{Appendix}
\appendix
\renewcommand{\thesection}{A\arabic{section}}

\section{Hand Module}
\label{sec:hand_module}
In this section, we provide more details of our stand-alone hand module described in Sec.~\ref{sec:hand_module_main}.
We first describe the hand mesh and hierarchy of the hand skeleton. Then we discuss how to preprocess data from different datasets.
Then we discuss the implementation details of our hand module.
In the end, we discuss the hand datasets used in our paper and provide experiment results to demonstrate the effectiveness of adopting different datasets.

\subsection{Hand Module Details}
\label{sec:hand_module_detail}

\begin{figure}[t!]
	\begin{center}
		\includegraphics[width=\linewidth]{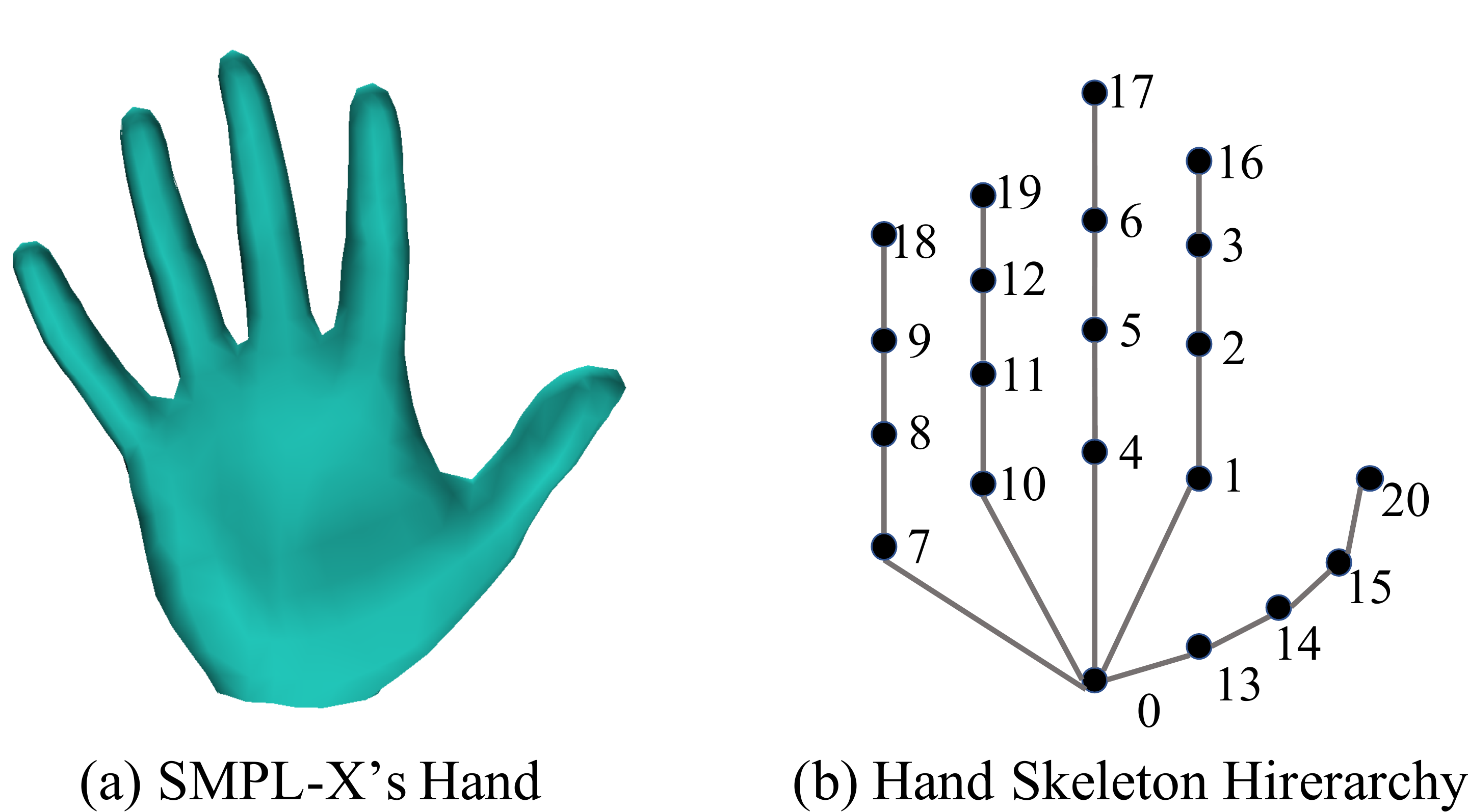}
	\end{center}
	\vskip -0.3cm
	\caption{\small We take the hand part of SMPL-X as a stand-alone hand model for hand pose estimation. The example mesh is shown in (a) and the skeleton hierarchy is shown in (b).}
	\label{fig:hand_visualize}
	\vspace{0.1cm}
\end{figure}

\newpara{Hand Mesh.}
The visualization of cropped hand mesh and hierarchy of the hand skeleton is depicted in Fig.~\ref{fig:hand_visualize}.

\newpara{Data preprocessing.}
Following the practice of 3D body pose estimation methods~\cite{kanazawa2018end, kanazawa2019learning, rong2019delving, kolotouros2019spin}, we include as many publicly available datasets as possible towards in-the-wild generalization ability.
To handle the discrepancy of annotations across different datasets, 
% To handle the discrepancy of annotations from different datasets,
%
we perform several pre-processing steps to make them consistent and compatible with our hand model.
The pre-processing includes:
(1) Rescaling all 3D keypoint annotations to be compatible with our hand model, by using the middle finger's knuckle length as a reference.
(2) Reordering the 3D keypoints to be the same as our hand model's skeleton hierarchy.

\newpara{Hand Module Implementation Details.}
Input images of the hand module are center-cropped surrounding the hands, where the bounding boxes for cropping are given by 2D hand keypoints. 
We use ground-truth 2D keypoints during training and predicted keypoints from OpenPose~\cite{cao2019openpose} during testing.
The cropped images are further padded and resized to $224 \times 224$.
During training, we apply data augmentations to each of the training images via random scaling, translation, rotation, color jittering, and synthetic motion blur.
The hand module architecture is based on ResNet-50~\cite{he2016deep} with two additional fully connected layers to map the output features of ResNet to vectors with $61$ dimension, which is composed of camera parameters $\boldsymbol{c}_h$ (3 dimensions), hand global rotation $\bphi_h$ (3 dimensions), hand pose parameters $\btheta_{h}$ (45 dimensions) and shape parameters $\bbeta_{h}$ (10 dimensions). 
The hand module is implemented with PyTorch~\cite{paszke2019pytorch}. The Adam optimizer~\cite{kingma2015adam} with learning rate $1\mathrm{e}{-4}$ is used to train the model. The hand module is trained until converge, which takes about $100$ epochs.

\begin{table}[t] \centering \small
	\caption{\small In training our hand module, we use five datasets including: FreiHAND~\cite{zimmermann2019freihand}, HO-3D~\cite{hampali2020honnotate}, MTC~\cite{xiang2019monocular}, STB~\cite{zhang20173d} and RHD~\cite{zimmermann2017learning}. We use MPII+NZSL~\cite{simon2017hand} for validation only.}
	\vskip -0.2cm
	\setlength\tabcolsep{2pt}
	\begin{tabular}{c|c|c|c|c}
		\hline
		Information $\rightarrow$    & \multirow{2}{*}{\# of Samples}  & \multirow{2}{*}{Pose Angles} &  \multirow{2}{*}{3D Joints} & \multirow{2}{*}{2D Joints} \\
		Dataset $\downarrow$		 & 	 		&         &    &  \\   	
		\hline
		FreiHAND~\cite{zimmermann2019freihand}             &  60K     & \cmark  &  \cmark  & \cmark  \\
		HO-3D~\cite{hampali2020honnotate}                &  60K     & \cmark  &  \cmark  & \cmark  \\
		MTC~\cite{xiang2019monocular}                  &  200K    &         &  \cmark  & \cmark \\
		STB~\cite{zhang20173d}                  &  20K     &         &  \cmark  & \cmark \\
		RHD~\cite{zimmermann2017learning}                  &  50K     &         &  \cmark  & \cmark \\
		MPII+NZSL~\cite{simon2017hand}            &  10K     &         &          & \cmark  \\
		\hline
	\end{tabular}	
	\label{tab:dataset_information}
	%\vspace{0.5 cm}
\end{table}

\begin{figure*}[t]
	\begin{center}
		\includegraphics[width=1.0\linewidth]{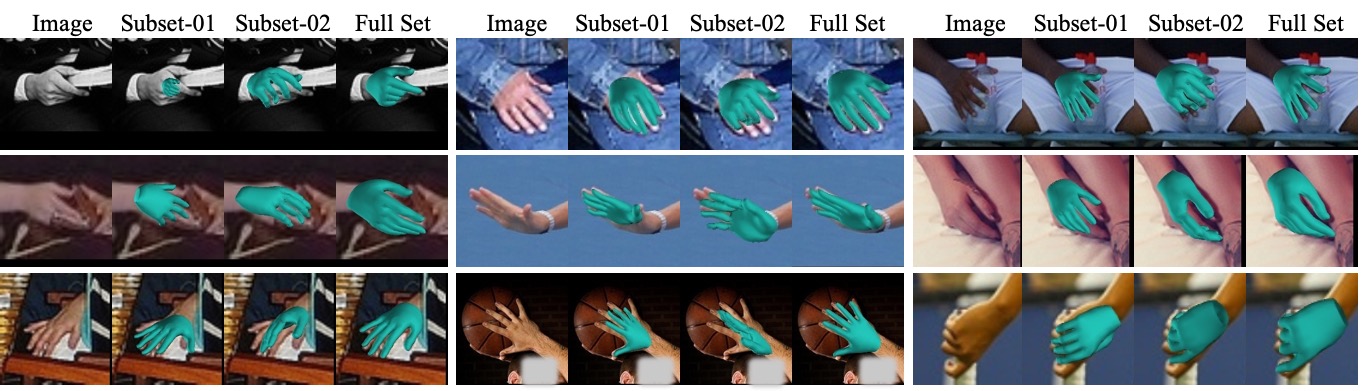}
	\end{center}
	\vskip -0.5cm
	\caption{\small Ablation study on training datasets. We show qualitative ablation study on using different datasets in training our hand model. ``Subset-01'' means using the combination of datasets FreiHADN~\cite{zimmermann2019freihand} and HO-3D~\cite{hampali2020honnotate}. ``Subset-02'' means using the combination of datasets: STB~\cite{zhang20173d}, RHD~\cite{zimmermann2017learning} and MTC~\cite{xiang2019monocular}. ``Full set'' means using all the above datasets. The images are selected from COCO dataset~\cite{lin2014microsoft}.}  
	\label{fig:ablation_dataset}
	\vspace{-0.2cm}
\end{figure*}

\subsection{Hand Datasets}
To increase our hand module's generalization ability, we adopt as many public datasets as possible. 
The summary of different datasets is listed in Tab.~\ref{tab:dataset_information}.
The details of these datasets are listed below.

\noindent \textbf{FreiHAND}.
FreiHAND~\cite{zimmermann2019freihand} is a dataset with ground truth 3D hand joints and MANO parameters for real human hand images. 
The 3D annotations are obtained by a multi-camera system and a semi-automated approach. The obtained data is further augmented with synthetic backgrounds.
In our experiments, we randomly select 80\% of samples from the original training set as training data and use the remaining 20\% of samples for validation.

\noindent \textbf{HO-3D}. 
HO-3D dataset~\cite{hampali2020honnotate} is a dataset aiming to study the interaction between hands and objects. 
The dataset has 3D joints and MANO pose parameters for hands. It also has 3D bounding boxes for objects the hands interact with. In this paper, we only use 3D annotations of hands.
The training set is composed of different sequences, each of which records one type of hand-object interaction.
Following the similar practice in processing FreiHAND, we randomly choose 80\% of sequences from the original training set as training data and use the remaining 20\% of sequences for validation.

\noindent \textbf{MTC.} 
Monocular Total Capture~\cite{xiang2019monocular} is a dataset captured by Panoptic Studio~\cite{joo2015panoptic,joo2017panoptic} in a multi-view setup with 30 HD cameras. It has 3D hand joints annotations for both body and hands. The sequences are mainly the range of motion data of multiple subjects.
To polish the dataset, we filter out erroneous samples where hands are not visible or too small. 

\noindent \textbf{STB.} 
Stereo Hand Pose Tracking Benchmark~\cite{zhang20173d} is composed of 15,000 training samples and 3,000 testing samples. 
During our experiments, we only use the RGB images and the paired 3D joint annotations.
We use the training set of STB to train our model and compare our models with other state-of-the-art methods on the validation set.
To unify the definition of joints, following the practice of~\cite{cai2018weakly,ge20193d}, we move the root joint from the palm center to the wrist.

\noindent \textbf{RHD.} 
Rendered Hand Dataset~\cite{zimmermann2017learning} is a synthetic dataset that has 2D and 3D hand joint annotations. 
It is composed of 41,258 training samples and 2,728 testing samples. 
We train our models on the training set and compare our models with other state-of-the-art methods on the testing set.

\noindent \textbf{MPII+NZSL.} 
MPII+NZSL dataset~\cite{simon2017hand} is composed of in-the-wild images with manually annotated 2D hand joints.
It includes challenging images with occlusion, blur, and low resolution. 
To show our models' generalization ability, we do not incorporate MPII+NZSL into training set. We only use it for validation.

\subsection{Ablation Study on Hand Datasets.}
\label{sec:ablation_hand_dataset}
\begin{table}[t] \centering \small %%\scriptsize
	%\rowcolors{1}{}{lightgray}

	\vskip -0.2cm
		\setlength\tabcolsep{2pt}
	\resizebox{\columnwidth}{!}{
		\begin{tabular}{c|c||c|c|c||c}
			\toprule 
			FreiHAND~\cite{zimmermann2019freihand}     & HO-3D~\cite{hampali2020honnotate}     & MTC~\cite{xiang2019monocular}      &   STB~\cite{zhang20173d}   & RHD~\cite{zimmermann2017learning}     & MPII+NZSL~\cite{simon2017hand} \\
			\midrule
			\cmark       &           &          &         &         & 0.482 \\
			& \cmark    &          &         &         & 0.367 \\
			\cmark       & \cmark    &          &         &         & 0.526 \\
			&           & \cmark   & \cmark  & \cmark  & 0.556 \\
			\cmark       & \cmark    & \cmark   &         &         & 0.595 \\
			\cmark       & \cmark    & \cmark   & \cmark  &         & 0.598 \\
			\cmark       & \cmark    & \cmark   &         & \cmark  & 0.645 \\
			\cmark       & \cmark    & \cmark   & \cmark  & \cmark  & 0.655 \\
			\bottomrule
		\end{tabular}	
	}
	%\vspace{0.2cm}
	\caption{\small Ablation study on dataset. We show the results of our hand module trained with different datasets. These models are evaluated on MPII+NZSL~\cite{simon2017hand} using 2D AUC as metric. For data augmentation, we use all the available datasets.}
	\setlength\tabcolsep{2pt}
	\label{tab:ablation_dataset}
\end{table}

We examine the influence of using different datasets in this subsection. The results are listed in Tab.~\ref{tab:ablation_dataset} and Fig.~\ref{fig:ablation_dataset}. As expected, the results in Tab.~\ref{tab:ablation_dataset} show that using more datasets will lead to better performance. 
We also show examples of qualitative comparison in Fig.~\ref{fig:ablation_dataset}. 
Similar to the conclusion from the quantitative study, the qualitative results show that incorporating more datasets can increase the models' generalization ability and generate more precise results for in-the-wild images.
In the figure, ``Subset-01'' means using the combination of datasets FreiHAND~\cite{zimmermann2019freihand} and HO-3D~\cite{hampali2020honnotate}.
``Subset-02'' means using the combination of datasets: STB~\cite{zhang20173d}, RHD~\cite{zimmermann2017learning} and MTC~\cite{xiang2019monocular}. 
``Full set'' means using all the datasets.

\begin{figure*}[t]
	\centering
	\includegraphics[width=0.85\linewidth]{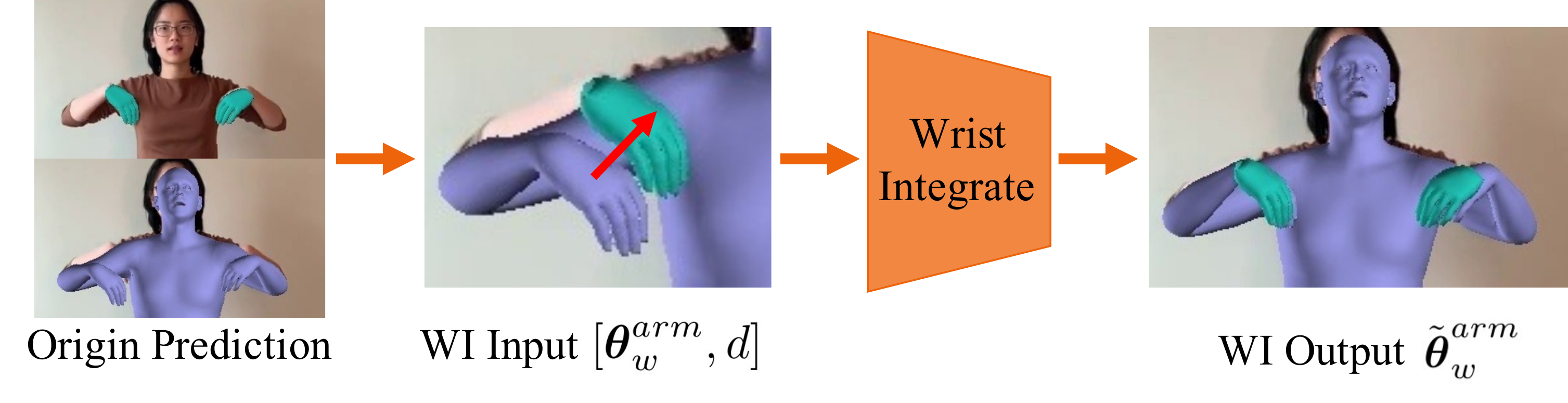}
	\vspace {-0.5cm}
	\caption{Overview of the proposed wrist integration network.}
	\label{fig:wi_network}
	\vspace {-0.2cm}
\end{figure*}

\section{Wrist Integration Network.}
\label{sec:wrist_integration_supp}

\subsection{Framework Details.}
The framework of wrist integration network is depicted in Fig.~\ref{fig:wi_network}.
Given hand prediction from hand module and whole-body prediction from copy-paste integration module, the role of wrist integration network is to adjust the arm poses to move hands to the locations determined by the hand module.
The input of the wrist integration network is composed of two parts.
The first part is the 2D directional vector $d$ that the arm needs to follow in the image space. 
It is obtained from the wrist locations calculated from the body module and those from the hand module, followed by a normalization with the length of arm.
The second part $\btheta_w^{arm}$ is the pose parameters of the elbow and shoulder joints the copy-paste integration. 
The original shoulder poses $\btheta_w^{s}$ and elbow poses $\btheta_w^{e}$ are all local rotations relative to their parent defined in the kinematics of body skeleton. 
To ease the training of the wrist integration network, we transform the shoulder poses to the global orientation $\bphi_w^{s}$, so that the wrist integration network does not need to consider other body parts' poses.
The elbow poses are kept as the local rotation.
In this way, the input arm poses $\btheta_w^{arm}$ is composed of global orientation of shoulder $\bphi_w^{s}$ and local rotations of elbow $\btheta_w^{e}$. 

\begin{equation}
\begin{gathered}
\btheta_w^{arm} = [\bphi_w^{s}, \btheta_w^{e}] = [\Upsilon_s(\btheta_w^{s}, \btheta_w), \btheta_w^{e}],
\label{eq:wi_input}
\end{gathered}
\end{equation}
where $\Upsilon_s$ is the function to convert the local shoulder rotation $\btheta_w^{s}$ to global shoulder orientation $\bphi_w^{s}$.
$\Upsilon_s$ can be implemented by forwarding the kinematics of the SMPL-X body skeleton from the root to the shoulder.

Taking the arm poses $\btheta_w^{arm}$ and 2D directional vector as input, the wrist integration network predicts the updated arm poses $\ddot{\btheta}_w^{arm}$.
The whole process is formulated as Eq.~\eqref{eq:full_wi}. 
\begin{equation}
    \begin{gathered}
		\ddot{\btheta}_w^{arm} = \mathcal{H}(\btheta_w^{arm}, \textbf{d})
	\label{eq:full_wi}
\end{gathered}
\end{equation}
It is worth nothing that $\ddot{\btheta}_w^{arm}$ obtained in this step is not the final updated arm poses $\tilde{\btheta}_w^{arm}$ defined in Eq.~\eqref{eq:wrist_integration}.
$\ddot{\btheta}_w^{arm}=[\tilde{\bphi}_w^{s}, \tilde{\btheta}_w^{e}]$ is composed of updated global shoulder orientation $\tilde{\bphi}_w^{s}$ and updated local elbow rotation $ \tilde{\btheta}_w^{e}$. 
To make the obtained shoulder poses compatible with the whole-body poses $\btheta_w$ defined in Eq.~\eqref{eq:copy_paste}, we convert the global shoulder orientation $\tilde{\bphi}_w^{s}$ back to the local shoulder rotation $\tilde{\btheta}_w^{s}$.
This process is similar to converting global hand orientation $\bphi_h$ to local hand rotation $\btheta_h$ defined in Eq.~\eqref{eq:copy_paste}.
Since the updated local elbow rotation $ \tilde{\btheta}_w^{e}$ is already the local rotation, there is no need to further process it.
The final updated arm poses $\tilde{\btheta}_w^{arm}$ with pose shoulder and elbow arms in the local rotation, as defined in Eq.~\ref{eq:wrist_integration}, can be formulated as:
\begin{equation}
\begin{gathered}
%\tilde{\btheta}_w^{s} = \Gamma_s(\bphi_s, \btheta_w) \\
\tilde{\btheta}_w^{arm} = [ \tilde{\btheta}_w^s, \tilde{\btheta}_w^e  ] = [  \Gamma_s(\bphi_s, \btheta_w), \tilde{\btheta}_e ]
\label{eq:wi_output}
\end{gathered}
\end{equation}
where $\Gamma_s$ is the function to convert the global shoulder orientation $\tilde{\bphi}^s_w$ to local shoulder rotation $\tilde{\btheta}_w^s$. Its definition is similar to $\Gamma_l$ and $\Gamma_r$ defined in Eq.~\eqref{eq:copy_paste}.

\subsection{Implementation Details.}
The wrist integration network is a MLP with 6 layers. 
The input dimension is $8$, which is composed of $\btheta_w^{arm} \in \mathbb{R}^{2\times3}$ and $d \in \mathbb{R}^{2}$.
The output dimension is $6$ which is $\tilde{\btheta}_w^{arm} \in \mathbb{R}^{2 \times 3}$.
The dimension of the intermediate layers are $128$, $256$, $512$, $256$, $128$, separately.
For wrist integration network, we assume the input arm poses and output arm poses are only right arm poses.
During training, we only use right arm poses and right hand directional movement.
During inference, the input initial left arm poses and 2D directional vector are vertically flipped as there were right arms and hands. 
The output arm poses are then flipped back to the left arm pose space.

\begin{figure}
	\centering
	\includegraphics[width=\linewidth]{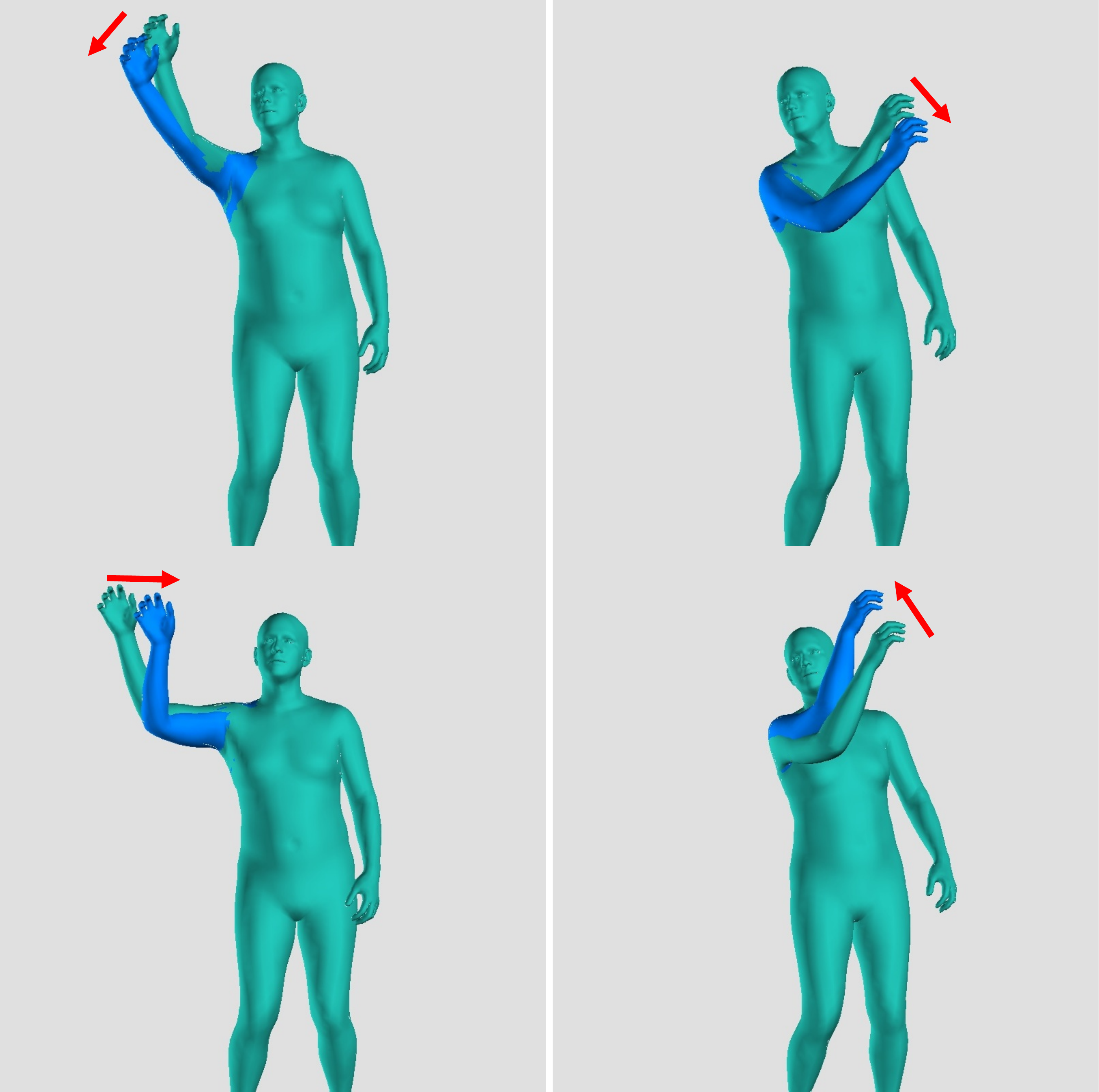}
	\vspace {-0.4cm}
	\caption{Samples of synthesized data for training wrist integration network. The~\textcolor{red}{right arrows} are the random translation $\dot{d}$ to be added.The~\textcolor{mygreen}{green bodies} are rendered from the original optimization output. The ~\textcolor{myblue}{blue arms} are the obtained arm poses $\tilde{\btheta}_w^{arm}$ after applying random translation.}
	\label{fig:wi_data}
\end{figure}

\subsection{Training Data.}
To train $\mathcal{H}$, we generate a synthetic dataset by capturing our own a range of motion of right arms. 
We first capture videos with varying arm poses.
Then we use the optimization method introduced in Section~\ref{sec:integration_main} to obtain the pseudo ground-truth 3D pose estimations for those video frames.
Suppose for one data sample, we obtain $\bphi_w$, $\btheta_w$,   $\bbeta_w$, $\boldsymbol{c}_w$, which are global rotation, whole-body pose, whole-body shapes and camera parameters.
To increase the model's generalization ability to varying body shapes, we add random noise to the obtain body shape $\bbeta_w$. %and obtain $\bar{\bbeta}_w$.
Then we use the updated body shapes and other obtained parameters to get projected right 2D hand joints $\boldsymbol{J}^{2D}_{rh}$ by projecting the predicted 3D right hand joints $\boldsymbol{J}^{3D}_{rh}$ using Eq.~\ref{eq:2d_projection}.
After that, we add random translation $\dot{\textbf{d}} = [\dot{d}_x, \dot{d}_y]$ to the original predicted 2D hand joints to obtain permutated 2D hand joints $\boldsymbol{\dot{J}}^{2D}_{rh} = \boldsymbol{J}^{2D}_{rh} + \dot{\textbf{d}}$.
We run the optimization method again to obtain the new poses using the permutated 2D hand joints $\boldsymbol{\dot{J}}^{2D}_{rh}$ in calculating $\mathcal{F}^{2d}$ defined in Eq.~\eqref{eq:full_opt}.
In this process, we fix other parameters and only update the right shoulder and right elbow poses, obtaining $\tilde{\btheta}_w^{arm}$.
In the end, we obtain training data with hand directional vector $\dot{\textbf{d}}$, original poses $\btheta_w^{arm}$ and updated arm poses $\tilde{\btheta}_w^{arm}$. They form the training data to train wrist integration network defined in Fig.~\ref{fig:wi_network}.

We visualize several generated data samples in Fig.~\ref{fig:wi_data}. 
The right arrows are the random translation $\dot{\textbf{d}}$ to be added.
The green bodies are rendered from the original optimization prediction. 
The blue arms are the obtained arm poses $\tilde{\btheta}_w^{arm}$ after applying randomly translating.

\section{Implementation Details}
\label{sec:implementation_detail}

\newpara{Bounding Bboxes.}
We use a lightweight Pytorch OpenPose implementation~\cite{osokin2018lightweight_openpose}, which only predicts 2D body keypoints, to obtain body bounding bboxes. 
Face and hand bounding bboxes can also be obtained from OpenPose~\cite{cao2019openpose}. 
For live demo requiring higher speed, the bounding boxes are obtained by projecting the face and hand part of the estimated 3D body to image space.

\begin{figure}[!t]
	\begin{center}
		\includegraphics[width=0.95\linewidth]{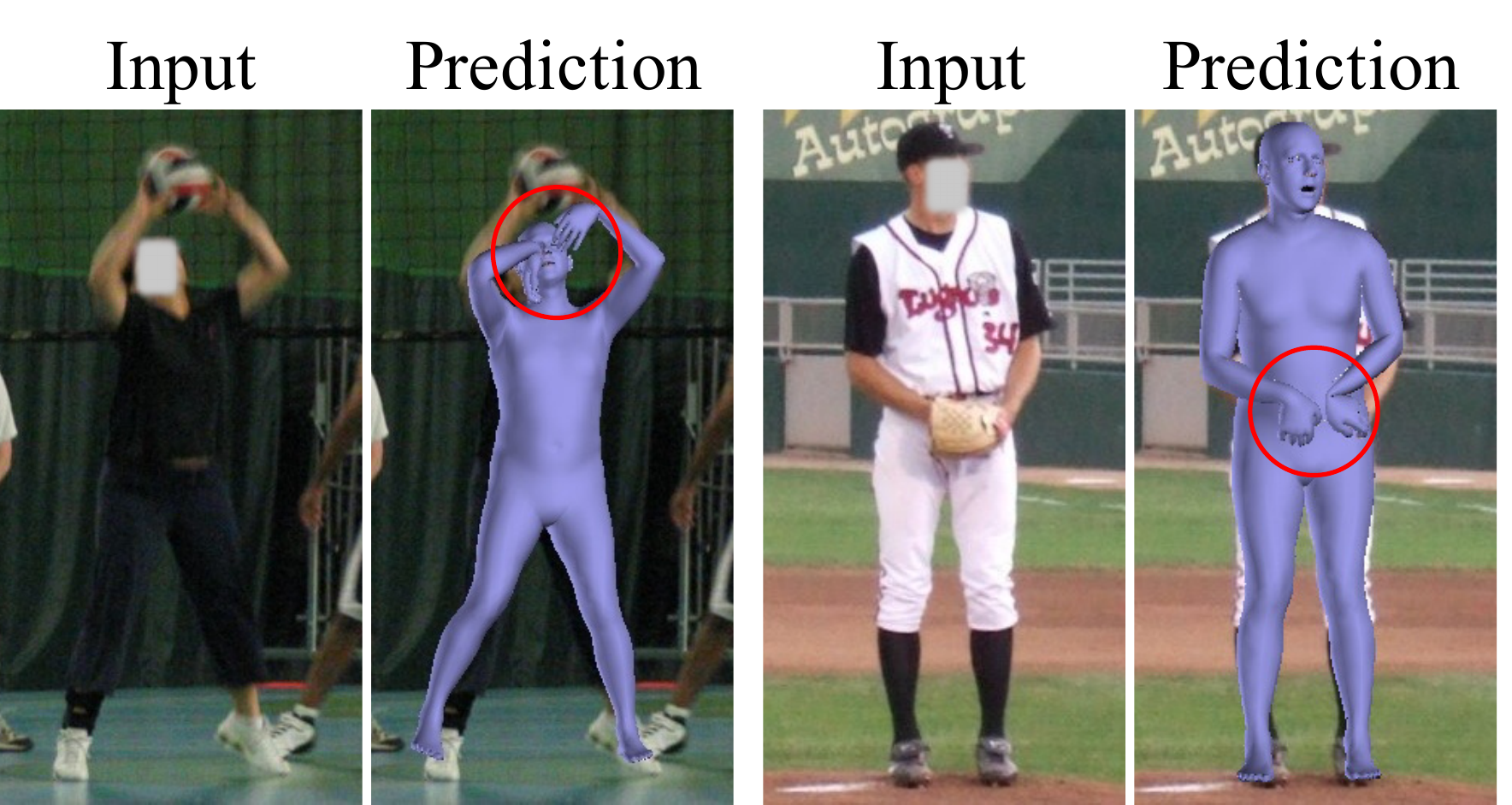}
	\end{center}
	\vskip -0.5 cm
	\caption{\small Typical failure cases from~\textit{FrankMocap}. It is shown that typical failure cases are caused by severe occlusion and blurry.}
	\label{fig:failure}
	%\vspace{-0.3 cm}
\end{figure}

\begin{figure}[!t]
	\begin{center}
		\includegraphics[width=0.95\linewidth]{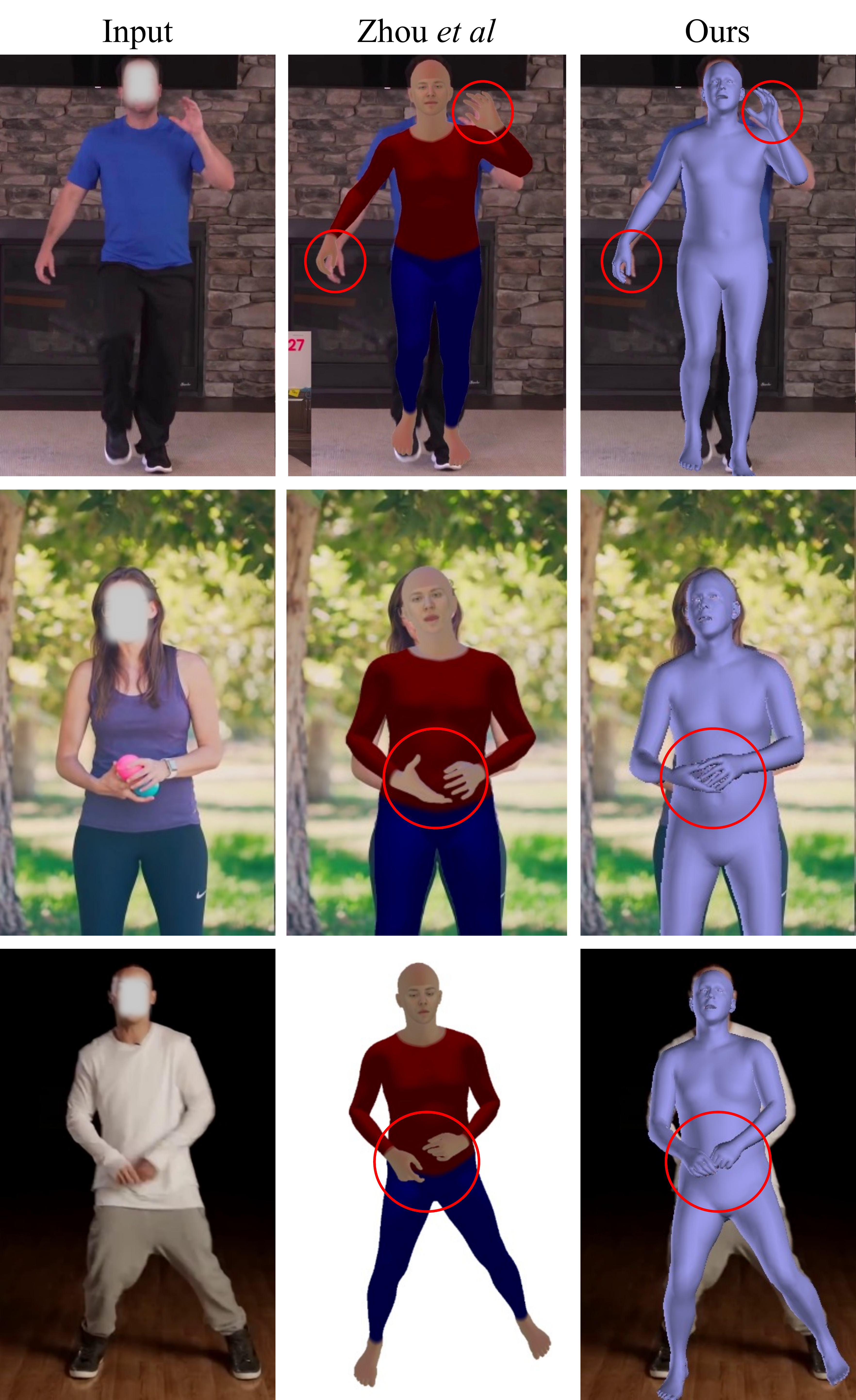}
	\end{center}
	\vskip -0.5 cm
	\caption{\small Quanlitative comparison with Zhou~\etal~\cite{zhou2021monocular}. Hand regions are marked with red circles. It is revealed that our method can generate more precise 3D hand poses than Zhou~\etal~\cite{zhou2021monocular} do.}
	\label{fig:compare_detnet}
	%\vspace{-0.3 cm}
\end{figure}

\section{Failure Cases}
We show several typical failure cases in Fig.~\ref{fig:failure}. It is revealed that failure cases are typically caused by severe occlusion and blurry.

\section{More Qualitative Results}
\label{sec:qualitative_supp}

\newpara{Comparison with Zhou~\etal.}
Firstly, we show qualitative comparisons with Zhou~\etal~\cite{zhou2021monocular} in Fig.~\ref{fig:compare_detnet}. It is worth noting that Zhou~\etal~\cite{zhou2021monocular} have not released their official code yet.
Therefore, we compare our methods with them using images shown in their paper.
The results demonstrate that our methods can generate more accurate hand and body poses than Zhou~\etal~\cite{zhou2021monocular} does.

\begin{figure*}
	\centering
	\includegraphics[width=0.95\linewidth]{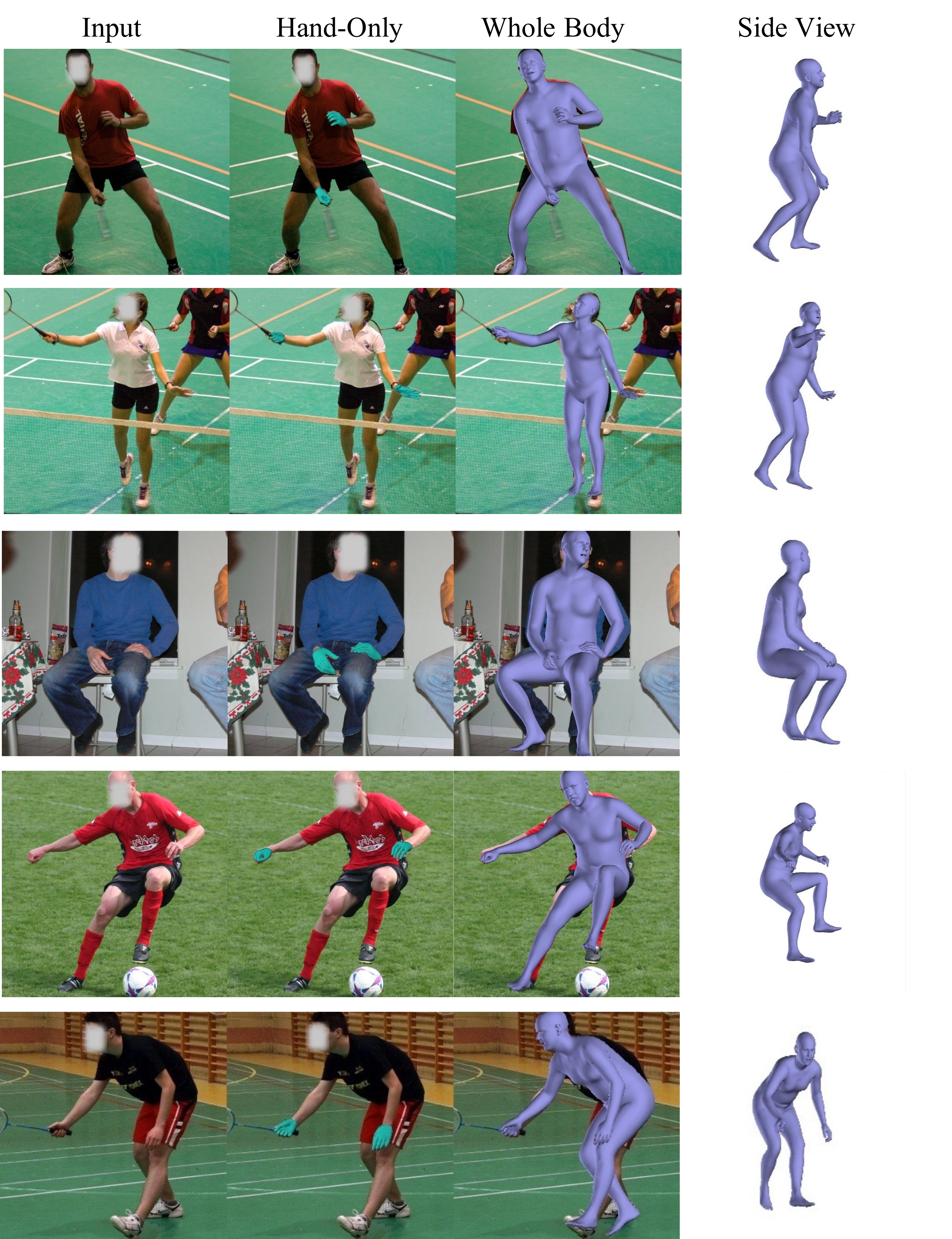}
	\caption{Qualitative results for \textit{FrankMocap} with wrist integration network adopted.}
	\label{fig:supp_qualitative}
\end{figure*}

\newpara{Demo.}
We show more qualitative results of \textit{FrankMocap} in Fig..~\ref{fig:supp_qualitative}. The integration module is the Wrist Integration Network.
More qualitative results are shown in the demo videos.

\clearpage

%\clearpage
%{\small
%\bibliographystyle{ieee_fullname}
%\bibliography{frank_mocap}
%}
%
%\end{document}

% reference
{\small
	\bibliographystyle{ieee_fullname}
	\bibliography{long,mocap}
}

\end{document}